\documentstyle[12pt]{article} 

\title{Dependence and Relevance: A probabilistic view}

\author{{\bf Dan Geiger}\\
Technion---Israel Institute of Technology\\
Haifa, Israel, 32000 \\
\and  
{\bf David Heckerman}\thanks{Current mailing address: One
Microsoft Way, 9S/1024, Redmond, WA 98052-6399}\\
Computer Science Department\\
University of California\\
Los Angeles, CA 90024}

\date{February 2, 1993}

\newtheorem{theorem}{Theorem}
\newenvironment{definition}{\vspace{1ex}\noindent\bf
Definition \begin{rm}}{\smallskip\end{rm}}

\newtheorem{lemma}[theorem]{Lemma}

\newcommand{\bmath}{\bf}

\newcommand{\vare}{e}
\newcommand{\e}{{\bf e}}
\newcommand{\ebar}{{\bf \bar{e}}}
\newcommand{\eone}{{\bf e'}}
\newcommand{\etwo}{{\bf \e''}}
\newcommand{\stare}{{\bf e^*}}

\newcommand{\Xuno}{{\bf X'}}
\newcommand{\Xdoz}{{\bf X''}}
\newcommand{\Yuno}{{\bf Y'}}
\newcommand{\Ydoz}{{\bf Y''}}

\newcommand{\Auno}{{\bf A'}}
\newcommand{\Adoz}{{\bf A''}}
\newcommand{\Buno}{{\bf B'}}
\newcommand{\Bdoz}{{\bf B''}}

\newcommand{\Aone}{{\bf A_1'}}
\newcommand{\Atwo}{{\bf A_2'}}
\newcommand{\Atre}{{\bf A_3'}}
\newcommand{\Afor}{{\bf A_4'}}

\newcommand{\sAone}{{\bf A_1^*}}
\newcommand{\sAtwo}{{\bf A_2^*}}
\newcommand{\sAtre}{{\bf A_3^*}}
\newcommand{\sAfor}{{\bf A_4^*}}

\newcommand{\tsAone}{{\bf A_1'}}
\newcommand{\tsAtwo}{{\bf A_2'}}
\newcommand{\tsAtre}{{\bf A_3'}}
\newcommand{\tsAfor}{{\bf A_4'}}

\newcommand{\tAone}{{\bf A_1''}}
\newcommand{\tAtwo}{{\bf A_2''}}
\newcommand{\tAtre}{{\bf A_3''}}
\newcommand{\tAfor}{{\bf A_4''}}

\newcommand{\Bone}{{\bf B_1'}}
\newcommand{\Btwo}{{\bf B_2'}}
\newcommand{\Btre}{{\bf B_3'}}
\newcommand{\Bfor}{{\bf B_4'}}

\newcommand{\sBone}{{\bf B_1^*}}
\newcommand{\sBtwo}{{\bf B_2^*}}
\newcommand{\sBtre}{{\bf B_3^*}}
\newcommand{\sBfor}{{\bf B_4^*}}

\newcommand{\starC}{{\bf C^*}}
\newcommand{\starD}{{\bf D^*}}

\long\def\comment#1{}

\begin{document}

\setlength{\baselineskip}{21pt}


\maketitle

\begin{abstract}
\noindent We examine three probabilistic 
concepts related to the sentence ``two variables
have no bearing on each other''.
We explore the relationships between these three concepts
and establish their relevance to the process of 
constructing similarity networks---a tool for acquiring 
probabilistic knowledge from human experts.
We also establish a precise relationship between
connectedness in Bayesian networks and relevance
in probability.
\end{abstract}

\section{Introduction}

The notion of relevance between pieces of information plays
a key role in the theory of
Bayesian networks and in the way they are used for inference.
The intuition that guides the construction of 
Bayesian networks draws from the analogy between 
``connectedness'' in graphical representations
and ``relevance'' in the domain represented,
that is, two nodes connected along some 
path correspond to variables of mutual relevance.

We examine three formal concepts related to the sentence 
``variables $a$ and $b$ have no bearing on each other''.
First, two variables $a$ and $b$
are said to be {\em mutually irrelevant} if
they are conditionally independent given any value of any subset of 
the other variables in the domain.
Second, two variables are said to be 
{\em uncoupled} if the set of variables representing the domain
can be partitioned into two independent sets
one containing $a$ and the other containing $b$.
Finally, two variables $a$
and $b$ are {\em unrelated} if the corresponding
nodes are disconnected in every minimal Bayesian network
representation (to be defined).

The three concepts, mutual-irrelevance, uncoupledness,
and unrelatedness are not identical.
We show that uncoupledness and 
unrelatedness are always equivalent but sometimes differ from
the notion of mutual-irrelevance.
We identify a class of models called {\em transitive} for which
all three concepts are equivalent.
Strictly positive binary
distributions (defined below) are examples of transitive models.
We also show that ``disconnectedness'' in graphical representations
and mutual-irrelevance in the domain represented
coincide for every transitive model and for none other.

These results have theoretical and practical ramifications.
Our analysis uses a qualitative
abstraction of conditional independence known
as graphoids [Pearl and Paz, 1989], and demonstrates the 
need for this abstraction in manipulating conditional 
independence assumptions 
(which are an integral part of every probabilistic reasoning engine). 
Our results also simplify
the process of acquiring probabilistic knowledge 
from domain experts via similarity networks.

This article is organized as follows:
A short overview on graphoids and their Bayesian network
representation is provided in Section 2.
(For more details consult Chapter 3 in Pearl, 1988.)
Section 3 and 4 investigate properties of
mutual-irrelevance, uncoupledness, and unrelatedness and
their relation to each other. Section 5 discusses two
definitions of similarity networks.  Section 6 shows that
for a large class of probability distributions these
definitions are equivalent.

\section{Graphoids and Bayesian Networks}

Since our definitions of mutual-irrelevance, uncoupledness, and
unrelatedness all rely on the notion of conditional independence,  it
is useful to abstract probability distributions to reflect this fact.
In particular, every probability distribution is viewed  as a list of
conditional independence statements with no reference to numerical
parameters. This abstraction, called a {\em graphoid}, was
proposed by Pearl and Paz [1989] and further discussed by Pearl [1988]
and Geiger [1990]. 

Throughout the discussion we consider a finite set
of variables $U = \{ u_1 ,\ldots, u_n \}$ each of 
which is associated with
a finite {\em set of values} $d( u_i)$ and
a probability distribution
$P$ having the Cartesian product
$\mbox{\large $\times$}_{u_i\in U} d(u_i)$
as its sample space.

\begin{definition}
A probability distribution
$P$ is {\em defined over $U$} if
its sample space is
$\mbox{\large $\times$}_{u_i\in U} d(u_i)$.
\end{definition}

We use lowercase letters possibly subscripted 
(e.g., $a$, $b$, $x$ or $u_i$)
to denote variables, and use uppercase letters 
(e.g., $X$, $Y$, or $Z$) to denote sets of variables.  A bold
lowercase or uppercase letter refers to a value (instance) 
of a variable or of a set of variables, respectively. 
A value {\boldmath $X$} of a set of variables $X$
is an element in the Cartesian product 
$\mbox{\large $\times$}_{x \in X} d(x)$,
where $d(x)$
is the set of values of $x$.  
The notation $X = {\bf X}$
stands for $x_1 = {\bf x_1} , \ldots , x_n = {\bf x_n}$,
where
$X= \{x_1 , \ldots , x_n \}$ and $\bf x_i$ is a value of $x_i$.

\begin{definition}
The expression $I(X,Y \mid Z)$, where $X$, $Y$, and $Z$ are
disjoint subsets of $U$, is called an {\em independence statement},
or {\em independency}.
Its negation $\neg I(X,Y \mid Z)$ is called 
a {\em dependence statement}, or {\em dependency}.
An independence or dependence statement is defined {\em over} 
$V \subseteq U$ if it mentions only elements of $V$.
\end{definition}

\begin{definition}
Let $U = \{ u_1 ,\ldots, u_n \}$ be a finite set
of variables with $d(u_i)$ and $P$ as above.
An independence
statement $I(X,Y \mid Z)$ is said to
{\em hold for} $P$ if
for every value
$\bmath X$, $\bmath Y$, and $\bmath Z$ of $X$, $Y$, and $Z$,
respectively
\begin{displaymath}
P(X = {\bf X} \mid  Y = {\bf Y} , Z = {\bf Z}) \;\;= \;\;
P(X = {\bf X} \mid  Z = {\bf Z})
\end{displaymath}
or $P(Z = {\bf Z}) = 0$.
Equivalently, $P$ is said to {\em satisfy} $I(X,Y \mid Z)$.
Otherwise, $P$ is said to {\em satisfy} $\neg I(X,Y \mid Z)$.
\end{definition}

\vspace{1ex}

\begin{definition}
When $I(X,Y \mid Z)$ holds for $P$, then $X$ and $Y$ are
{\em conditionally independent} 
relative to $P$ and if $Z = \emptyset$, 
then $X$ and $Y$ are {\em marginally independent} relative to $P$.
\end{definition} 

Every probability distribution defines a {\em dependency model}.

\begin{definition}{\cite{bk:pearl}}
A {\em dependency model} $M$
over a finite set of elements $U$ is a set of triplets
$(X, Y \mid  Z)$, where $X$, $Y$ and $Z$ are disjoint subsets
of $U$. 
\end{definition}

The definition of dependency models does not assume any structure
on the elements of $U$. Namely, an element of $U$ could be, for
example, a node in some graph or a name of a variable. In
particular,
if each $u_i \in U$ is associated with a finite set
$d(u_i)$, then every probability distribution having
the Cartesian product $\mbox{\large $\times$}_{u_i\in U} d(u_i)$
as it sample space defines a dependency model via
the rule:
\begin{equation}
(X,Y \mid Z) \in M \mbox{  if and only if  } 
I(X,Y \mid Z) \mbox{ holds in $P$},
\label{rule}
\end{equation}
for every disjoint subsets $X, Y,$ and $Z$ of $U$.
Dependency models constructed using Equation~\ref{rule}
have some interesting properties that
are summarized in the definition below.

\begin{definition}{}
A {\em Graphoid} $M$
over a finite set $U$ is any set of triplets
$(X, Y \mid Z)$, where $X$, $Y$, and $Z$ are disjoint subsets
of $U$ such that
the following axioms are satisfied:

\vspace{1ex}

\noindent
Trivial Independence
\begin{equation}
( X,  \emptyset \mid Z)\in M \label{trivial}
\end{equation}
Symmetry
\begin{eqnarray}
( X, Y \mid Z )\in M  & \Rightarrow& (Y , X \mid  Z )\in M \label{symmetry}
\end{eqnarray}
Decomposition
\begin{eqnarray}
( X, Y  \cup  W \mid  Z )\in M &\Rightarrow& 
( X, Y \mid Z)\in M \label{decomposition}
\end{eqnarray}
Weak union
\begin{eqnarray}
(X, Y\cup W \mid Z ) \in M &\Rightarrow&  ( X, Y \mid Z\cup  W ) \in M 
\label{weak}
\end{eqnarray}
Contraction
\begin{eqnarray}
\nonumber
\lefteqn{(X, Y \mid  Z)\in M \: \& \: ( X, W \mid Z \cup Y ) \in M \;
\Rightarrow} \hspace{1in} \\
& & ( X, Y \cup W \mid Z ) \in M.
\label{contraction}
\end{eqnarray}
The above relations are called 
the {\em graphoid axioms}.\footnote{This 
definition differs slightly from that given  
in \cite{graphoid}, where axioms (\ref{symmetry}) 
through (\ref{contraction})
define semi-graphoids. A variant of these axioms
was first studied by David [1979]\nocite{dawid79} and 
Spohn [1980]\nocite{spohn80}.}
\end{definition}

\vspace{1ex}

Using the definition of
conditional independence, it is easy to show that 
each probability distribution defines a graphoid 
via Equation~\ref{rule} [Pearl, 1988]. The graphoid
axioms have an appealing interpretation.
For example, the weak union axiom states: If $Y$ and $W$
are conditionally independent of $X$, given a knowledge base
$Z$, then $Y$ is conditionally independent of $X$ given
$W$ is added to the known knowledge base $Z$.
In other words,
the fact that a piece of information $W$, which is
conditionally independent of $X$, becomes known, does
not change the status of $Y$; $Y$ remains conditionally
independent of $X$ given the new {\em irrelevant} 
information $W$ \cite{bk:pearl}.

Graphoids are suited to represent the qualitative
part of a task that requires a probabilistic analysis.
For example, suppose an alarm system is installed 
in your house in order to detect burglaries; and suppose
it can be activated by two separate sensors.  
Suppose also that, when the alarm sound is activated,
there is a good chance that a police patrol will show up.
We are interested in computing the probability of
a burglary given a police car is near your house.

The dependencies in this story can be represented by a graphoid.
We consider five
binary variables, {\em burglary}, {\em sensorA}, {\em sensorB},
{\em alarm}, and {\em patrol},
each having two values {\em yes} and {\em no}. 
We know that
the outcome of the two sensors are conditionally independent given
{\em burglary}, and that 
{\em alarm} is conditionally independent of 
{\em burglary} given the outcome of the sensors.
We also know that {\em patrol\/} is conditionally independent of
{\em burglary} given {\em alarm}. (Assuming that only the alarm
prompts a police patrol.) This qualitative information
implies that the following three triplets
must be included in a dependency model that
describes the above story:
({\em sensorA}, {\em sensorB} $\mid$  {\em burglary}),
({\em alarm}, {\em burglary} $\mid$  \{{\em sensorA}, {\em sensorB}\})
and 
({\em patrol}, 
\{{\em burglary}, {\em sensorA}, {\em sensorB}\} $\mid$ {\em alarm}).

The explicit representation of all triplets of a dependency model
is often impractical, because there are an exponential number of 
possible triplets.  Consequently, an implicit representation
is needed.  We will next describe such a representation.

\begin{definition}{\cite{bk:pearl}}
Let $M$ be a graphoid over $U$. 
A directed acyclic graph
$D$ is a {\em Bayesian network of $M$} if
$D$ is constructed from $M$
by the following steps: assign a {\em construction order\/}
$u_1, u_2, ,\ldots, u_n$ to the elements in
$U$, and designate a node for each $u_i$. 
For each $u_i$ in $U$, 
identify a set
$ \pi ( u_i ) \subseteq \{ u_1 ,\ldots, u_{i-1} \}$ such that 
\begin{equation}
\label{graphoid_eq}
(\{u_i\} , \{u_1, \ldots , u_{i-1}\} \setminus \pi(u_i) \mid  \pi(u_i)) \in M.
\end{equation}
Assign a link from every element in $\pi ( u_i )$ to $u_i$. 
The resulting network is {\em minimal} if, for each $u \in U$,
no proper subset of $\pi(u)$ satisfies Equation~(\ref{graphoid_eq}).
\end{definition}


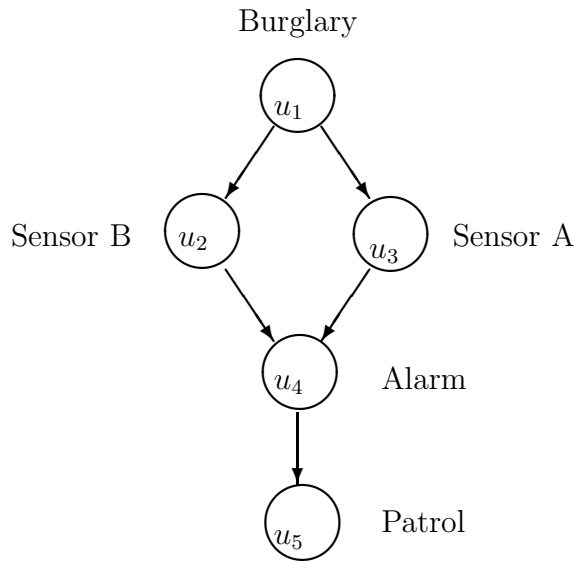
\begin{figure}
\begin{center}

\setlength{\unitlength}{0.0125in}%
\begin{picture}(185,234)(190,493)
\thicklines
\put(312,509){\circle{32}}
\put(311,572){\circle{32}}
\put(349,630){\circle{32}}
\put(270,631){\circle{32}}
\put(310,688){\circle{32}}
\put(320,675){\vector( 2,-3){ 20}}
\put(310,555){\vector( 0,-1){ 30}}
\put(340,615){\vector(-2,-3){ 20}}
\put(280,615){\vector( 2,-3){ 20}}
\put(300,675){\vector(-2,-3){ 20}}
\put(285,715){\makebox(0,0)[lb]{\raisebox{0pt}[0pt][0pt]{Burglary}}}
\put(375,625){\makebox(0,0)[lb]{\raisebox{0pt}[0pt][0pt]{Sensor A}}}
\put(190,625){\makebox(0,0)[lb]{\raisebox{0pt}[0pt][0pt]{Sensor B}}}
\put(345,565){\makebox(0,0)[lb]{\raisebox{0pt}[0pt][0pt]{Alarm}}}
\put(345,505){\makebox(0,0)[lb]{\raisebox{0pt}[0pt][0pt]{Patrol}}}
\put(300,680){\makebox(0,0)[lb]{\raisebox{0pt}[0pt][0pt]{$u_1$}}}
\put(340,620){\makebox(0,0)[lb]{\raisebox{0pt}[0pt][0pt]{$u_3$}}}
\put(260,625){\makebox(0,0)[lb]{\raisebox{0pt}[0pt][0pt]{$u_2$}}}
\put(300,565){\makebox(0,0)[lb]{\raisebox{0pt}[0pt][0pt]{$u_4$}}}
\put(300,500){\makebox(0,0)[lb]{\raisebox{0pt}[0pt][0pt]{$u_5$}}}
\end{picture}

\caption{An example of a Bayesian network}
\end{center}
\label{fig1}
\end{figure}


A Bayesian network of the burglary story is shown
in Figure~\ref{fig1}. We shall see that aside
of the triplets that were used to construct
the network, the triplet 
({\em patrol}, {\em burglary} $\mid$ \{{\em sensorA}, {\em sensorB}\})
follows from the topology of the network.
In the remainder of this section, we describe a general methodology
to determine which triplets
of a graphoid $M$ are represented in a Bayesian network of $M$.
The criteria of {\em d}-separation, defined below, provides the answer.
Some preliminary definitions are needed.

\begin{definition}
The {\em underlying graph} of a Bayesian network is an 
undirected graph obtained from the Bayesian network by
replacing every link with an undirected edge.
\end{definition}

\begin{definition}
A {\em trail} in a Bayesian network is a sequence of links that form
a simple (cycle-free) path in the underlying graph.
Two nodes are {\em connected} in a Bayesian network
if there exists a trail connecting them. 
Otherwise they are {\em disconnected}.  If $x \rightarrow y$
is a link in a Bayesian network, then $x$ is a {\em parent}
of $y$ and $y$ is a {\em child} of $x$. If there is a
directed path of length greater than zero from $x$ to $y$,
then $x$ is an {\em ancestor} 
of $y$ and $y$ is a {\em descendant} of $x$.
\end{definition}

\begin{definition}{\em (Pearl, 1988)}
A node $b$ is called a
{\em head-to-head} node wrt (with respect to) 
a trail $t$ if there are two
consecutive links $a \rightarrow b$ and $b \leftarrow c$ on $t$. 
\end{definition}

For example, $u_2 \rightarrow u_4 \leftarrow u_3$ is a trail
in Figure~\ref{fig1} and $u_4$ is a head-to-head node 
with respect to this trail.

\begin{definition}{\em (Pearl, 1988)}
A trail $t$ is 
{\em active wrt} a set of nodes $Z$ if (1)
every head-to-head node wrt $t$
either is in $Z$ or has a descendant in $Z$ and (2) every other node
along $t$ is outside $Z$.  Otherwise, the trail is said to be 
{\em blocked} (or {\em d-separated}) by $Z$.
\end{definition}

In Figure~\ref{fig1}, for example, both trails
between $\{u_2\}$ and $\{u_3\}$ are d-separated by $Z= \{u_1\}$; 
the trail
$u_2 \leftarrow u_1 \rightarrow u_3$ is d-separated by $Z$ 
because 
node $u_1$, which is not a head-to-head node wrt
this trail, is in $Z$.
The trail $u_2 \rightarrow u_4 \leftarrow u_3$ is 
d-separated by $Z$,
because node $u_4$ and its descendant $u_5$
are outside $Z$.  In contrast, $u_2 \rightarrow u_4 \leftarrow u_3$
is not {\em d}-separated by $Z'= \{u_1, u_5\}$ because $u_5$ is
in $Z'$.

The theorem below is the major 
building block for most of the developments
presented in this article.

\begin{theorem}{\cite{soundness}}
Let $D$ be a Bayesian network of a graphoid $M$ over $U$; and
let $X$, $Y$, and $Z$ be three disjoint subsets of $U$.
If all trails between a node
in $X$ and a node in $Y$ are {\em d}-separated by $Z$,
then $(X, Y \mid Z) \in M$.
\label{graphoid_th}
\end{theorem}

For example, 
in the Bayesian network of Figure~\ref{fig1},
all trails between $u_1$ and $u_5$ are {\em d}-separated by 
$\{u_2, u_4\}$.  Thus, Theorem~\ref{graphoid_th} guarantees that
$(u_5, u_1 \mid \{u_2,u_3\}) \in M$, where $M$ is
the graphoid from which this network was constructed.
\footnote{Within the expression of independence statements,
we often write $u_i$ instead of $\{u_i\}$.}
Furthermore, Geiger and Pearl [1990]
show that no other graphical criteria reveals more triplets
of $M$ than does {\em d}-separation.

Geiger et al.~[1990]
generalize Theorem~\ref{graphoid_th} to networks that include 
{\em deterministic nodes} (i.e., nodes whose value is a {\em
function} of their parents' values).
Shachter [1990] obtains related results.
Lauritzen et al. [1990] establish another graphical criteria and show
that it is equivalent to {\em d}-separation.

\section{Three Notions of Relevance}

We can now define mutual-irrelevance,
uncoupledness, and unrelatedness, and study their properties.

\begin{definition} Let $M$ be a graphoid over
$U$, and let $x, y \in U$. 
\begin{itemize}
\item
$x$ and $y$ are {\em uncoupled} if
there exist a partition $U_1, U_2$ of $U$ such that
$x \in  U_1 $, $y \in U_2$, and 
$(U_1, U_2 \mid \emptyset) \in M$. 
Otherwise, $x$ and $y$ are {\em coupled}, 
denoted {\em coupled(x,y)}.

\item
$x$ and $y$ are {\em unrelated} if
$x$ and $y$ are disconnected in every minimal
Bayesian network of $M$. Otherwise, $x$ and $y$ are
{\em related}, denoted {\em related(x,y)}.

\item
$x$ and $y$ are {\em mutually irrelevant} if
$(x, y \mid Z) \in M$ for every $Z \subseteq U \setminus \{x,y\}$.
Otherwise, $x$ and $y$ are {\em mutually-relevant}, 
denoted {\em relevant(x,y)}.

\end{itemize}
\end{definition}

We could have defined these concepts
using conditional independence.  By choosing
the graphoid framework, however, we gain in two
aspects.  First, we emphasize that all the properties
that we discuss are proved using the graphoid
axioms.  We do not use any properties of probability
theory that are not summarized in these axioms.
Second, our results are more general in that they
appeal to any graphoid, not necessarily a graphoid
that is defined by conditional independence,
or even a graphoid defined by a probability distribution. 
Examples of other types of graphoids are given in
Pearl [1988].

Later in this section, we show that
if two nodes $x$ and $y$ are disconnected in one minimal 
Bayesian network 
of $M$, then $x$ and $y$ are disconnected in every minimal 
network of $M$.
Thus, to check whether $x$ and $y$ are unrelated, it
suffices to examine whether or not they are connected in one minimal 
network representation rather than examine all possible 
minimal networks.
This observation, demonstrated by Theorem~\ref{theorem_3},
offers a considerable reduction in complexity. Based on the
development that leads to Theorem~\ref{theorem_3}, we also prove
that $x$ and $y$ are unrelated if and only if they are uncoupled. 

\begin{definition}
A {\em connected component} $C$ of a Bayesian network $D$ is
a subgraph of $D$ in which every two nodes are connected (by a trail).
A connected component is {\em maximal} 
if there exists no proper 
super-graph of $C$ that is a connected component of $D$.
\end{definition}

\begin{lemma}
Let $D$ be a Bayesian network of a graphoid $M$ over $U$,
and let $A$ and $B$ be subsets of $U$.
If all nodes in $A$ are disconnected from all nodes in 
$B$, then
$(A, B \mid  \emptyset ) \in M$.
\label{last_lemma}
\end{lemma}

\noindent {\bf Proof:}  
There is no active trail between a node in $A$ and a node in $B$.
Thus, by Theorem~\ref{graphoid_th}, 
$(A, B \mid  \emptyset) \in M$.  $\Box$

\begin{lemma}
Let $D$ be a Bayesian network of a graphoid $M$ over $U$,
$x$ be in $U$, $Z$ be the set of $x$'s parents, 
and $Y$ be the set of all nodes that are not descendants 
of $x$ except $x$'s parents.
Then, $(x, Y \mid Z) \in M$.
\label{ordering_lemma}
\end{lemma}

\noindent {\bf Proof:}  
The set $Z$\/ d-separates all trails between a node in $Y$ and $x$,
because each such trail either passes through a parent of $x$
and therefore is blocked by $Z$, or each such trail must reach $x$ through
one of $x$'s children and thus must have a head-to-head node $w$, where
neither $w$ nor its descendants are in $Z$.
Thus, by Theorem~\ref{graphoid_th}, $(x, Y \mid Z) \in M$. $\Box$

\begin{lemma}
Let $D$ be a minimal Bayesian network of a graphoid $M$ over $U$,
$x$ be in $U$, and $Z_1 \cup Z_2$ be $x$'s parents.
Then $(x, Z_1 \mid Z_2) \not\in M$, unless $Z_1 = \emptyset$.
\label{parenthood}
\end{lemma}

\noindent {\bf Proof:} Since $Z_1 \cup Z_2$ are the parents of $x$, by
Lemma~\ref{ordering_lemma}, $(x, Y \mid Z_1 \cup Z_2) \in M$, where $Y$
is the set of $x$'s non-descendants except its parents.
Assume, by contradiction, that $(x, Z_1 \mid Z_2) \in M$
and $Z_1 \neq \emptyset$.
The two triplets imply by the contraction axiom that
$(x, Z_1 \cup Y \mid Z_2) \in M$.
Since $Z_1 \neq  \emptyset$, $Z_2$ is a proper subset
of $Z_1 \cup Z_2$, where $Z_1 \cup Z_2$ are the parents of $x$ in $D$. 
Hence $D$ is not minimal, because Equation~\ref{graphoid_eq}
is satisfied by a proper subset of $x$'s parents---a
contradiction.
$\Box$

\begin{definition}
A set $\{X_1, X_2\}$ is a {\em partition} of $X$ iff
$X= X_1 \cup X_2$, $X_1 \cap X_2 = \emptyset$,  
$X_1 \neq \emptyset$, and 
$X_2 \neq \emptyset$.
\end{definition}

\begin{lemma}
Let $M$ be a graphoid over $U$, 
$D$ be a minimal Bayesian network of $M$,
and $D_X$ be a
connected component of $D$ with a set of nodes $X$.
Then, there exists no partition $X_1, X_2$ of $X$ such that
$(X_1, X_2 \mid \emptyset) \in M$.
\label{more_lemma}
\end{lemma}

\noindent {\bf Proof:}  Suppose $X_1, X_2$ is a partition of $X$ such
that $(X_1, X_2 \mid \emptyset) \in M$. Since $X_1$ and $X_2$ are
connected in $D$, there must exist a link between a node in $X_1$ and 
a node in $X_2$. Without loss of generality, assume it is directed
from a node in $X_1$ to a node $u$ in $X_2$.

Let $Z_1$, $Z_2$
be the parents of $u$ in $X_1$ and $X_2$, respectively. The
triplet $(X_1, X_2 \mid \emptyset)$, which we assumed to
be in $M$,
implies---using symmetry and decomposition---that 
$(u \cup Z_2,Z_1 \mid \emptyset)$ is in $M$.
By symmetry and weak union,
$(u ,Z_1 \mid Z_2 )$ is in $M$ as well.
Thus, by Lemma~\ref{parenthood}, the network $D$ is not minimal,
unless $Z_1 = \emptyset$.  We have assumed, however, that $u$
has a parent in $X_1$.
Hence, $Z_1 \neq \emptyset$.
Therefore, $D$ is not minimal, contrary to our assumption.
$\Box$

\begin{theorem}
\label{theorem_3}
Let $M$ be a graphoid over $U$.
If two elements of $U$ are
disconnected in some minimal Bayesian network of $M$,
then they are disconnected in every minimal Bayesian network 
of $M$.
\end{theorem}

\comment{
{\bf Proof} Assume $a$ and $b$ are disconnected in one minimal network
of $M$. Let $U_1$ denote the variables
connected to $a$ and $U_2$ denote the rest. By Lemma~\ref{last_lemma},
$(U_1, U_2 \mid \emptyset) \in M$. Thus, $a$ and $b$ are uncoupled.
By Lemma~\ref{more_lemma}, they are disconnected 
in every minimal network of $M$. $\Box$
}

\noindent {\bf Proof:} It suffices to show that any two minimal
Bayesian networks
of $M$ share the same maximal connected components.
Let $D_A$ and $D_B$ be 
two minimal Bayesian networks of $M$.  Let $C_A$ 
and $C_B$ be maximal connected components
of $D_A$ and $D_B$, respectively.  Let $A$ 
and $B$ be the nodes of $C_A$ 
and $C_B$, respectively.  We 
show that either $A = B$ or $A \cap B =  \emptyset$.  This
demonstration will complete the proof, because for an arbitrary 
maximal connected component $C_A$ in $D_A$ there must exist
a maximal connected component in $D_B$ that shares at least one node
with $C_A$.  Thus, by the above claim, it must have 
exactly the same nodes as $C_A$.  Therefore, each maximal connected component 
of $D_A$ shares the same nodes with exactly one maximal connected
component of $D_B$.  Hence, $D_A$ and $D_B$ share the same
maximal connected components.

Since $D_A$ is a minimal Bayesian network of $M$ and
$C_A$ is a maximal connected component of $D_A$, 
by Lemma~\ref{last_lemma},
$(A , U \setminus A \mid \emptyset ) \in M$.
Using symmetry and decomposition, 
$( A \cap B  , B \setminus A \mid  \emptyset ) \in M$.  
Thus, by Lemma~\ref{more_lemma}, for $C_B$ to be a maximal 
connected component,
either $A \cap B $ or $B \setminus A$ must be empty, lest $D_B$ would
not be minimal.
Similarly, for $C_A$ to be a maximal connected component, $A \cap B$ 
or $A \setminus B$ must be empty.
Thus, either $A = B$ or $A \cap B =  \emptyset$.  $\Box$

\vspace{1ex}

\begin{theorem}
\label{theorem_4}
Two variables $x$ and $y$ of a graphoid $M$ over $U$ are unrelated
iff they are uncoupled.
\end{theorem}

\noindent {\bf Proof:} If $x$ and $y$ are unrelated, then let
$U_1$ be the variables connected to $x$ in some minimal network of $M$,
and $U_2$ be the rest of the variables in $U$. 
By Lemma~\ref{last_lemma}, $(U_1, U_2 \mid \emptyset) \in M$.
Thus, $x$ and $y$ are uncoupled. 

If $x$ and $y$ are uncoupled, then
there exist a partition $U_1, U_2$ of $U$ such that
$x \in  U_1 $, $y \in U_2$, and 
$(U_1, U_2 \mid \emptyset) \in M$. 
We show that in every minimal
Bayesian network $D$ of $M$, nodes $x$ and $y$ do not reside in the same
connected component.  Thus, $x$ and $y$ are unrelated. 
Assume, to the contrary, that $x$ and $y$ reside in the same 
maximal connected component of a minimal Bayesian network $D$ of $M$,
and that $X$ are the nodes in that component.
The statement $(U_1 \cap X, U_2 \cap X \mid\emptyset)$ follows
from $(U_1, U_2 \mid \emptyset) \in M$ by the symmetry and
decomposition axioms. 
Moreover, $U_1 \cap X$ and $U_2 \cap X$ are not empty, because
they include $x$ and $y$, respectively. Since $U_1$ and $U_2$
are disjoint, the two sets
$U_1 \cap X$, $U_2 \cap X$ partition $X$.
Therefore, by Lemma~\ref{more_lemma},
$D$ cannot be minimal, contrary to our assumption.
$\Box$

Theorem~\ref{theorem_4} shows that $x$ and $y$ are related if and only
if they are coupled. 

\section{Transitive Graphoids}

In this section, we show that if $x$ and $y$ are coupled, then
$x$ and $y$ are mutually-relevant. Then, we identify conditions
under which the converse holds, and provide
an example in which these conditions are not met.

\begin{theorem}
Let $M$ be a graphoid over $U$. Then,
for every $x,y \in U$,
\[
relevant(x,y) \Rightarrow coupled(x,y).
\]
\label{needed}
\end{theorem}

\noindent {\bf Proof:}  Suppose $x$ and $y$ are not coupled. Let $U_1$,
$U_2$ be a partition of $U$ such that $x \in U_1$,
$y \in U_2$ and $(U_1, U_2 \mid \emptyset) \in M$.
We show that $x$ and $y$ must be mutually irrelevant. 
Let $Z$ be an arbitrary subset of $U \setminus \{x,y\}$. 
Let $Z_1 = Z \cap U_1$ and 
$Z_2 = Z \cap U_2$.
The statement $(U_1, U_2 \mid \emptyset) \in M$ implies---by
decomposition and symmetry axioms---that $(\{x\} \cup Z_1, \{y\} \cup
Z_2 \mid \emptyset) \in M$.
By symmetry and weak union, $(x,y \mid Z_1 \cup Z_2) \in M$.
Thus, $(x,y \mid Z) \in M$  for every $Z \subseteq U \setminus\{x, y\}$.
Hence, $x$ and $y$ are mutually irrelevant.
$\Box$

\vspace{1ex}

The converse of Theorem~\ref{needed} does not hold in general;
if $x$ and $y$ are mutually irrelevant, it does not imply 
that $x$ and $y$ are uncoupled.
For example, assume $M$ is a graphoid over $U=\{x,y,z\}$ that
consists of $(x, y \mid \emptyset ),\,(x, y \mid z )$ and the statements
implied from them by the graphoid axioms.
Then, $x$ and $y$ are mutually irrelevant, yet $x$ and $y$ are
coupled because $(x,\{y,z\} \mid \emptyset) \not\in M$ and
$(\{x,z\},y \mid \emptyset) \not\in M$.

To see that there is a probability distribution that induces this
graphoid, 
suppose $x$ and $y$ are the outcomes of two independent fair coins.
In addition, suppose that $z$ is a variable whose domain is 
$\{head,tail \} \times  \{ head, tail \}$
and whose value is $( i, j )$ if and only if the outcome of $x$ is $i$
and the outcome of $y$ is $j$.
Then $x$ and $y$ are mutually irrelevant, because $x$ and $y$ are 
marginally independent and independent given $z$.  Nevertheless, they 
are coupled, because neither $I(x,\{y,z\} \mid \emptyset)$ nor
$I(\{x,z\},y \mid \emptyset)$ hold for $P$.

A necessary and sufficient condition for the converse
of Theorem~\ref{needed} to hold, as we shall see, is
that the graphoid $M$ is transitive.

\begin{definition} 
A graphoid $M$ over $U$ is {\em transitive} if
for every $x,y,z \in M$,
\begin{equation}
{\em relevant}(x,y) \;\&\;{\em relevant}(y,z) \Rightarrow
{\em relevant}(x,z).
\end{equation}
\end{definition}

First, we show that transitivity is necessary.

\begin{theorem}
Let $M$ be a graphoid over $U$ such that for
every $x,y \in U$, coupled(x,y) implies relevant(x,y).
Then $M$ is a transitive graphoid.
\label{equivalence1}
\end{theorem}

\noindent {\bf Proof:}  
By Lemma~\ref{needed}, {\em relevant(x,y)} if and only if
{\em coupled(x, y)}.
Also, by Theorem~\ref{theorem_4},
{\em coupled(x, y)} if and only if {\em related(x, y)}.
Since {\em related} is a transitive relation, so is 
{\em relevant}. Thus, $M$ is transitive. $\Box$

Some preliminaries are needed before we show that transitivity
is a sufficient condition as well.

\begin{definition}
Let $M$ be a graphoid over $U$, 
and $A$, $B$ be two disjoint subsets of $U$.
Then $A$ and $B$ are {\em mutually irrelevant},
if $(A, B \mid Z) \in M$
for every $Z$ that is a subset of $U \setminus A \cup  B$.
\end{definition}

\begin{lemma}
\label{lemma_2}
Let $M$ be a graphoid over $U$, and $A$, $B$, and $C$ be
three disjoint subsets of $U$.  If $A$ and $B$ are 
mutually irrelevant, and $A$ and $C$ are mutually irrelevant,
then $A$ and $B \cup C$ are mutually irrelevant as well.
\label{composition_lemma}
\end{lemma}

\noindent {\bf Proof:} Denote 
the sentence ``$X$ and $Y$ are mutually irrelevant''
with $J(X, Y)$.  By definition, $J(A, B)$ 
implies $(A, B \mid  Z) \in  M$ and
$J(A, C)$ implies $(A, C \mid  Z \cup  B ) \in  M$,
where $Z$ is an arbitrary subset of $U \setminus A \cup B \cup C$.
Together, these
statements imply by the contraction axiom that
$(A, B \cup C \mid Z) \in M$.
Since $Z$ is arbitrary, $J(A, B \cup  C)$ holds.  $\Box$

\vspace{1ex}

As is well known from probability theory, 
if $A$ and $B$ are independent,
and $A$ and $C$ are independent, then, contrary to
our intuition, $A$ is not necessarily independent
of $B \cup C$.  Lemma~\ref{composition_lemma},
on the other hand, shows that
if $A$ and $B$ are mutually irrelevant, 
and $A$ and $C$ are mutually irrelevant,
then $A$ and $B \cup C$ must also be mutually irrelevant.

\begin{theorem}
\label{equivalence2}
If $M$ is a transitive graphoid over $U$, then
for every $x,y \in U$,
\[
coupled(x,y) \Rightarrow relevant(x,y).
\]
\end{theorem}

\noindent {\bf Proof:}  
Let $M$ be a transitive graphoid over $U$, and
$x$, $y$ be two arbitrary elements in $U$ such
that $x$ and $y$ are mutually irrelevant.  
We will show by induction on $|U|$
that if {\em relevant} is transitive, then
there exists a Bayesian network $D$ of $M$
where $x$ and $y$ are disconnected.  Consequently, $x$ and $y$ are
uncoupled (Theorem \ref{theorem_4}).

We construct $D$ in the ordering
$u_1 ( \equiv x) , u_2  (\equiv y) , u_3 ,\ldots, u_{n-1}, u_n 
(\equiv e)$ of $U$.  Assume $n = 2$.  Variables $x$ and $y$ are
mutually irrelevant.
Thus, $(x, y \mid \emptyset  ) \in M$.  Hence, $x$ and $y$ are not 
connected.  Otherwise, $n > 2$.  

Let $M_e$ be a dependency model over $U \setminus \{e\}$
formed from $M$ by removing all triplets involving $e$.
The model $M_e$ is a graphoid, because whenever
the left hand side of one of the graphoid axioms does not
mention $e$, then neither does the right hand side.
Let $D_e$ be a minimal Bayesian network of $M_e$
formed from $M_e$ by the construction order $u_1 ,\ldots, u_{n-1}$. 
Let $A$ be the set of nodes connected to $x$, let $B$ be the
set of nodes connected to $y$, and let $C$ be the
rest of the nodes in $D_e$.  The Bayesian network $D$ 
of $M$ is formed
from $D_e$ by adding the last node $e$ as a sink and
letting its parents be a minimal set that makes $e$ 
independent of all the rest of the variables
in $U$ (following the definition of minimal Bayesian networks).  

Since $x$ and $y$ are mutually irrelevant in $M$, it follows 
that they are also mutually irrelevant in $M_e$.
Thus, by the induction hypothesis,
$x$ and $y$ are disconnected in $D_e$.  
After node $e$ is added, a trail through $e$ might exists 
in $D$ that connects a node in $A$ and a node in $B$.  
We will show that there is none; if the
parent set of $e$ is minimal, then
either $e$ has no parents in $A$ or it has no parents in $B$, 
rendering $x$ and $y$ disconnected in $D$.

Since $x$ and $y$ are mutually irrelevant,
it follows that either $x$ and $e$ are mutually irrelevant or
$y$ and $e$ are mutually irrelevant, lest
$M$ would not be transitive. 
Without loss of generality, assume that $x$ and $e$ are
mutually irrelevant.
Let $x'$ be an arbitrary node in $A$.  
By transitivity it follows that either
$x$ and $x'$ are mutually irrelevant or $e$ and $x'$ are
mutually irrelevant, lest $x$ and $e$ would not be
mutually irrelevant, contrary to our selection of $x$.  If $x$
and $x'$ are mutually irrelevant, then by the induction hypothesis,
$A$ can be partitioned into two marginally 
independent subsets.  Thus, by Lemma~\ref{more_lemma}, $A$ would
not be connected in the Bayesian network $D_e$, 
contradicting our selection of $A$.  Thus,
every element $x' \in A$ and $e$ are mutually irrelevant.  
It follows that the entire set $A$ and $e$ are mutually irrelevant 
(Lemma \ref{composition_lemma}). Thus, in particular, 
$(e, A \mid \hat{B} \cup \hat{C}) \in M$, where 
$\hat{B}$ are the parents of $e$ in $B$, and
$\hat{C}$ are the parents of $e$ in $C$. 
Assume $\hat{A}$ is the set of parents of $e$ in $A$.
By decomposition, $(e, A \mid  \hat{B} \cup \hat{C}) \in M$ implies
$(e, \hat{A} \mid  \hat{B} \cup \hat{C}) \in M$.
By Theorem~\ref{parenthood}, $D$ is not minimal,
unless $\hat{A}$ is empty.
$\Box$

Theorems \ref{needed}, \ref{equivalence1} and \ref{equivalence2} 
show that the relations
{\em coupled} and {\em relevant} are identical for every
transitive graphoid and for none other. We emphasize that
these results apply also to every probability distribution
that defines a transitive graphoid. In section 6, we show
that many probability distributions indeed define transitive graphoids.
First, however, we pause to demonstrate the relationship of these
results to knowledge acquisition and knowledge representation.

\section{Similarity Networks}

Similarity networks were invented by Heckerman [1990] as
a tool for constructing large Bayesian networks from domain
experts judgements.
Heckerman used them to construct a large diagnosis
system for lymph-node pathology. 
The main advantage of similarity networks is their ability
to utilize statements of conditional independence
that are not represented 
in a Bayesian network, in order to reduce
more drastically the number of 
parameters a domain expert needs to specify. 
Furthermore, the construction of a large Bayesian network
is divided into several stages each of which involves
the construction of a small local Bayesian network. 
This divide and conquer approach helps
to elicit reliable expert judgements.
At the diagnosis stage, the local networks
are combined into one global Bayesian network that represents
the entire domain.  

In [Geiger and Heckerman, 1993], we
show how to use the local networks directly for inference
without converting them to a global Bayesian network,
and remove several technical restrictions imposed by the
original development. Also, we develop two simple definitions
of similarity networks which we present here informally.
In the next section, we show that although the two
definitions are conceptually distinct they often coincide.

A Bayesian network
of a probability distribution $P(u_1,\ldots, u_n)$
is constructed as defined
in Section 2 with an important addition. After the topology
of the network is set, we
also associate with each node a conditional
probability distribution: $P(u_i \mid \pi(u_i))$.
By the chaining rule it follows that 
\[ 
P(u_1, \ldots, u_n) = \prod P(u_i \mid u_1, \ldots, u_{i-1}) 
\]
and by the definition of 
$I(\{u_i\} , \{u_1, \ldots , u_{i-1}\} 
\setminus \pi(u_i) \mid  \pi(u_i))$
we further obtain
\[ 
P(u_1, \ldots, u_n) = \prod P(u_i \mid \pi(u_i)) 
\]
Thus, the joint distribution is represented by the network
and can be used for computing the posterior probability of
every variable, given a specific value for some other variables.
For example, for the network of the burglary story (Figure~\ref{fig1}),
we need to specify the following conditional
distributions:
$P({\em burglary})$,
$P({\em sensorA} \mid {\em burglary})$,
$P({\em sensorB} \mid {\em burglary})$,
$P({\em alarm} \mid  {\em sensorA}, {\em sensorB})$, and
$P({\em patrol} \mid {\em alarm})$. From these numbers,
we can now compute any probability involving these variables.

A similarity network is a set of Bayesian networks,
called the local networks, each constructed under 
a different set of hypotheses $H_i$.  In each local network
$D_i$, only those variables that ``help to distinguish''
between the hypotheses in $H_i$ are depicted. 
The success of this model stems from
the fact that only a small portion of variables helps
to distinguish between the carefully chosen set of hypotheses
$H_i$,$i=1 \ldots k$. Thus,
the model usually includes several small networks instead
of one large Bayesian network.

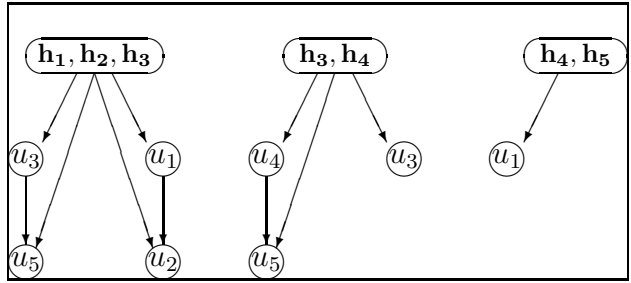
\begin{figure}
\setlength{\unitlength}{13pt}
\begin{center}
\begin{picture}(18,8)
\put(0,0){\framebox(18,8)}

\put(2.5,6.5){\makebox(0,0){\footnotesize $\bf h_1, h_2, h_3$}}
\put(2.5,6.5){\oval(4,1)}
\put(2,6){\vector(-1,-2){1}}
\put(2.5,6){\vector(-1,-3){1.7}}
\put(2.5,6){\vector(1,-3){1.7}}
\put(3,6){\vector(1,-2){1}}
\put(.5,3.5){\makebox(0,0){$u_3$}}
\put(.5,3.5){\circle{1}}
\put(.5,3){\vector(0,-1){2}}
\put(.5,.5){\makebox(0,0){$u_5$}}
\put(.5,.5){\circle{1}}
\put(4.5,3.5){\makebox(0,0){$u_1$}}
\put(4.5,3.5){\circle{1}}
\put(4.5,3){\vector(0,-1){2}}
\put(4.5,.5){\makebox(0,0){$u_2$}}
\put(4.5,.5){\circle{1}}
\put(9.5,6.5){\makebox(0,0){\footnotesize $\bf h_3, h_4$}}
\put(9.5,6.5){\oval(3,1)}
\put(9,6){\vector(-1,-2){1}}
\put(9.5,6){\vector(-1,-3){1.7}}
\put(10,6){\vector(1,-2){1}}
\put(7.5,3.5){\makebox(0,0){$u_4$}}
\put(7.5,3.5){\circle{1}}
\put(7.5,3){\vector(0,-1){2}}
\put(7.5,.5){\makebox(0,0){$u_5$}}
\put(7.5,.5){\circle{1}}
\put(11.5,3.5){\makebox(0,0){$u_3$}}
\put(11.5,3.5){\circle{1}}

\put(16.5,6.5){\makebox(0,0){\footnotesize $\bf h_4, h_5$}}
\put(16.5,6.5){\oval(3,1)}
\put(16,6){\vector(-1,-2){1}}
\put(14.5,3.5){\makebox(0,0){$u_1$}}
\put(14.5,3.5){\circle{1}}

\end{picture}
\end{center}
\caption{A similarity network}
\label{f2_fig}
\end{figure}

For example, Figure~\ref{f2_fig} is an example of a similarity
network representation of $P(h,u_1, \ldots, u_5)$ where $h$
is a distinguished variable that represents
five hypotheses $\bf h_1, \ldots, h_5$. In this similarity network,
variable $u_1$ is the only one that helps to discriminate
between $\bf h_4$ and $\bf h_5$, and variable 
$u_4$ is the only variable that
does not help to discriminate among $\{{\bf h_1, h_2, h_3}\}$. 

At the heart of the definition of similarity networks lies the
notion of discrimination.
The study of the relations {\em coupled}, {\em related}
and {\em relevant} presented in the previous sections,
enables us to formulate this notion in two ways, yielding
two types of similarity networks.

\begin{definition}{\cite{geigerheckerman93}}
A  similarity network constructed by including
in each local network $D_i$ every variable $x$, such that
$x$ and $h$ are {\em related} given that $h$ draws its values 
from $H_i$, is of {\em type 1}.
A  similarity network constructed by including
in each local network $D_i$ every variable $x$, such that
$x$ and $h$ are {\em relevant} given that $h$ draws its values
from $H_i$, is of {\em type 2}.
\end{definition}

In [Geiger and Heckerman, 1993], we show that type 1
similarity networks are diagnostically complete.  That is,
although some variables are removed from
each local network,
the posterior probability of every hypothesis, given any value
combination for the variables in $U$, can still be computed.
This result is reassuring because it guarantees that the 
computation we strive to achieve---namely, the 
computation of the posterior probability of the hypothesis---
can be performed.  The caveat of this result is that 
a knowledge engineer uses a type 1
similarity network to 
determine whether a variable  ``helps to discriminate''
the values in $H_i$, by asking a domain expert whether the node
corresponding to this variable
is connected to $h$ in the local Bayesian network
associated with $H_i$. This query might be too hard for
a domain expert to 
answer, because a domain expert does not necessarily
understand the properties of Bayesian networks.

On the other hand, a knowledge engineer uses a type 2 similarity
networks to determine whether a node ``does not help to discriminate''
the values in $H_i$, by asking an expert whether this variable can ever
help to distinguish the values of $h$, given that $h$ draws its
values from $H_i$.  This query concerns the subject matter of 
the domain; and
therefore a domain expert can more reliably answer the query. In fact,
this is the actual query Heckerman used in constructing his lymph-node
pathology diagnosis system.

Next, we show that these two definitions coincide for
large families of probability distributions.

\section{Transitive Distributions}

We show that
the relation {\em relevant} is transitive
whenever it is defined by a probability distribution
that belongs to one of the following two families:
strictly positive binary distributions and 
regular Gaussian distributions.
Hence, for these two classes of distributions,
type 1 and type 2 similarity networks are
identical.  Currently, we are working to show
that transitivity holds for other families.

\begin{definition}
A {\em strictly positive binary} distribution is a probability
distribution where every variable  has a domain of two values---say,
$0$ and $1$---and every combination of the variables' values has a
probability greater than zero.
A {\em regular Gaussian} distribution is a
multivariate normal distribution with finite nonzero variances
and with finite means.
\end{definition}

\vspace{1ex}

\begin{theorem}
Let $P(u_1, \ldots, u_n, e)$ be a
strictly positive binary distribution
or a regular Gaussian distribution.
Let $\{X_1, X_2\}$, $\{Y_1, Y_2\}$ and $\{Z_1, Z_2\}$
be three partitions of $U = \{u_1, \ldots, u_n\}$.
Let $R_1$ be $X_1 \cap Y_1 \cap Z_1$, and $R_2$ be
$X_2 \cap Y_2 \cap Z_2$.
Then,
\begin{eqnarray}
\lefteqn{
\nonumber
I(X_1, X_2 \mid \emptyset) \> \> \& \> \>
I(Y_1, Y_2 \mid \vare = \eone) \> \> \& \> \>
I(Z_1, Z_2 \mid \vare = \etwo) \Rightarrow} \\
& & I(R_1 , \{\vare\} \cup U \setminus R_1 \mid \emptyset) 
\> \vee \> 
I(R_2 , \{\vare\} \cup U \setminus R_2 \mid \emptyset)
\label{clean}
\end{eqnarray}
where $\eone$ and $\etwo$ are two distinct values of $e$.
\label{hardtheorem}
\end{theorem}

\hspace{-5ex}

When all three partitions are identical,
the above theorem can be phrased as follows.
If two sets of variables $A$ and $B$ are
marginally independent, and if
$I(A,B \mid e)$ holds as well, then
either $A$ is marginally independent of $\{e\} \cup B$ or
$B$ is marginally independent of $\{e\} \cup A$.
This special case has been stated in the literature
\cite{dawid79,bk:pearl}.

The proof of Theorem~\ref{hardtheorem} is given in Appendices A and B.
Theorem~\ref{hardtheorem} and the theorem below state
together that
strictly positive binary distributions 
and regular Gaussian distributions are transitive. 
Our assumptions of strict positiveness and
regularity were added to obtain a simpler proof.
We conjecture that both theorems still hold
when these restrictions are omitted.

\begin{theorem} 
Every probability distribution that satisfies
Equation \ref{clean} is transitive.
\label{maintheorem}
\end{theorem}

\noindent {\bf Proof:}
Let $P(u_1, \ldots, u_{n+1})$ be a probability distribution, let
$U = \{ u_1 , \ldots, u_{n+1} \}$; and
let $x$, $y$ be two arbitrary variables in $U$ such
that $x$ and $y$ are mutually irrelevant.  
We will show by induction on $|U|$
that if $P$ satisfies Equation~\ref{clean}, then 
$x$ and $y$ are uncoupled. 
Thus, according to Theorem~\ref{equivalence1},
$P$ is transitive. 

If $n = 1$, then the variables $x$ and $y$ are mutually irrelevant.
Thus, $I(x, y \mid  \emptyset)$ holds for $P$.  
Consequently, $x$ and $y$ are uncoupled.
Otherwise, assume without loss of generality
that $x$ is $u_1$ and $y$ is $u_2$,
and denote $u_{n+1}$ by $e$. Since $x$ and $y$ are mutually irrelevant
with respect to $P(u_1,\ldots, u_{n+1})$, $x$ and $y$ are
also mutually irrelevant with respect to $P(u_1,\ldots, u_n)$,
$P(u_1,\ldots, u_n \mid \vare = \eone)$, and
$P(u_1,\ldots, u_n \mid \vare = \etwo)$,
where $\eone$ and $\etwo$ are two distinct values of $u_{n+1}$.
Thus, by applying the the induction hypothesis three
times, we conclude that there are three partitions
$\{X_1, X_2\}$, $\{Y_1, Y_2\}$, and $\{Z_1, Z_2\}$
of $U = \{u_1, \ldots, u_n\}$ such that $x$ is in $X_1$,
$Y_1$, and $Z_1$, and $y$ is in $X_2$,
$Y_2$, and $Z_2$.  Hence, the antecedents of
Equation~\ref{clean} are satisfied.  Consequently,
$\{u_1, \ldots, u_{n+1}\}$ can be partitioned into two marginally
independent sets: either $R_1$ and $U \setminus R_1$, or
$R_2$ and $U \setminus R_2$, where 
$R_1$ is $X_1 \cap Y_1 \cap Z_1$ and $R_2$ is
$X_2 \cap Y_2 \cap Z_2$.
Because, in both cases,
one set contains $x$ and the other contains $y$,
it follows that $x$ and $y$ are uncoupled.
$\Box$


The practical ramification of this theorem is that
our concern of how to define discrimination via the relation
{\em related\/} or via {\em relevant} is not critical.
In many situations the two concepts coincide.

From a mathematical point of view, our proof demonstrates
that using an abstraction of conditional
independence---namely, the trinary relation $I$ combined with
a set of axioms---we are able to prove properties of very
distinct classes of distributions: strictly positive binary
distributions and regular Gaussian distributions. 


\section{Summary}

We have examined the notion of unrelatedness of variables
in a probabilistic framework, introduced three formulations
of this notion,
and explored their interrelationships.
From a practical point of view, these results legitimize
prevailing decomposition techniques of knowledge acquisition.
These results permit an expert to 
decompose the construction of a complex Bayesian network
into a set of Bayesian networks of manageable size.  

Our proofs use the qualitative notion of independence as
captured by the axioms of graphoids.  These proofs would have been 
harder to obtain had we used the usual
definitions of conditional independence.
This axiomatic approach enables us to identify a common
property---Equation~\ref{clean}---shared by two distinct
classes of probability distributions (regular Gaussian
and strictly positive binary),
and to use this property without attending to the detailed
characteristics of these classes.


\section*{Acknowledgments}
An earlier version of this article has
been presented in the sixth conference
on uncertainty in artificial intelligence
\cite{geigerheckerman90}.

\pagebreak[4]

\section*{Appendix A: Strictly Positive Binary Distributions}

Below, we prove Theorem~\ref{hardtheorem} for
strictly positive binary distributions.
First, we phrase the theorem differently.

\begin{theorem}
Strictly positive binary distributions 
satisfy the following 
axiom:\footnote{In complicated 
expressions, $A_1 A_2$ is used as a shorthand
notation for $A_1 \cup  A_2$ and $e A_1$ denotes
$\{ e \} \cup A_1$.}

\smallskip

$\!\!\!\!\!I( A_1 A_2 A_3 A_4 ,  B_1 B_2 B_3 B_4 \mid \emptyset )
\;\;\&\;$ \hfill \\
\hspace*{0.5cm}$I( A_1 A_2 B_3 B_4 ,  B_1 B_2 A_3 A_4 \mid  
\vare = \eone )  \;\;\&\; $\hfill \\
\hspace*{0.8cm}$ I( A_1 A_3 B_2 B_4 ,  B_1 B_3 A_2 A_4 \mid  
\vare = \etwo) \Rightarrow$ \hfill \\
\vspace{-0.3cm}
\begin{equation}
\!\!I( A_1 , \vare A_2 A_3 A_4 B_1 B_2 B_3 B_4 \mid \emptyset )   \,\vee 
 I( B_1 , \vare  A_1 A_2 A_3 A_4 B_2 B_3 B_4 \mid  \emptyset )
\label{pt-bin}
\end{equation}
where all sets mentioned are pairwise disjoint and do not contain
$e$, and {\boldmath $e'$} and {\boldmath $e''$}
are distinct values of $e$.
\end{theorem}

\comment{
\noindent
{\bf Theorem \ref{hardtheorem}}
{\em
Strictly positive binary distributions 
satisfy the following axiom:
\footnote{Throughout the appendices,
$XY$ is used as a shorthand
notation for $X \cup  Y$ and $e X$ denotes
$\{ e \} \cup X$.}

$\!\!\!\!\!I( A_1 A_2 A_3 A_4 ,  B_1 B_2 B_3 B_4 \mid \emptyset )
\;\;\&\;$ \hfill ($I_1$)\\
\hspace*{0.5cm}$I( A_1 A_2 B_3 B_4 ,  B_1 B_2 A_3 A_4 \mid  
\vare = \eone )  \;\;\&\; $\hfill ($I_2$)\\
\hspace*{0.8cm}$ I( A_1 A_3 B_2 B_4 ,  B_1 B_3 A_2 A_4 \mid  
\vare = \etwo) \Rightarrow$ \hfill ($I_3$)\\
\vspace{-0.3cm}
\begin{equation}
\!\!I( A_1 , \vare A_2 A_3 A_4 B_1 B_2 B_3 B_4 \mid \emptyset )   \,\vee 
 I( B_1 , \vare  A_1 A_2 A_3 A_4 B_2 B_3 B_4 \mid  \emptyset )
\label{pt-bin}
\end{equation}
where all sets mentioned are pairwise disjoint and do not contain
$e$, and {\boldmath $e'$} and {\boldmath $e''$}
are distinct values of $e$.
}

}

To obtain the original theorem, we set $A_1 A_2 A_3 A_4$,  $B_1 B_2
B_3 B_4$,  $A_1 A_2 B_3 B_4$, $B_1 B_2 A_3 A_4$,  $A_1 A_3 B_2 B_4$,
and  $B_1 B_3 A_2 A_4$ to be equal to $X_1, X_2, Y_1, Y_2, Z_1, and,
Z_2$ of the original theorem, respectively.

Denote the three antecedents of Equation~\ref{pt-bin} by
$I_1$, $I_2$, and $I_3$.  We need the following two Lemmas.

\begin{lemma} \label{lemBayes}
Let $X$ and $Y$ be two disjoint sets of variables, and let
$\e$ be an instance of a single binary 
variable $\vare$ not in $X \cup Y$.
Let $P$ be a probability distribution over the variables
$X \cup Y \cup \{\vare\}$.
If $I(X, Y \mid \vare=\e)$ holds for $P$, then for every pair of
instances $\Xuno , \Xdoz$ of $X$ and $\Yuno , \Ydoz $ of $Y$,
the following equation must hold:
\begin{displaymath}
\frac{P(\e |\Xuno \Yuno) P(\Xuno \Yuno)}
     {P(\e |\Xdoz \Yuno) P(\Xdoz\Yuno )} = 
\frac{P(\e | \Xuno \Ydoz) P(\Xuno \Ydoz )}
     {P(\e | \Xdoz \Ydoz) P(\Xdoz \Ydoz)}
\end{displaymath}
\end{lemma}
\noindent {\bf Proof:} Bayes' theorem states that
\begin{displaymath}
P(\Xuno | \e  \Yuno ) = 
\frac{P(\e | \Xuno  \Yuno) P(\Xuno \Yuno )}{P(\e \Yuno)}
\end{displaymath}
Thus,
\begin{displaymath}
\frac{P(\e | \Xuno \Yuno) P(\Xuno \Yuno )}
     {P(\e | \Xdoz  \Yuno) P(\Xdoz \Yuno )} \!= \!
\frac{P(\Xuno | \e , \Yuno )}{P(\Xdoz | \e , \Yuno )}  \!= \!
\frac{P(\Xuno | \e , \Ydoz )}{P(\Xdoz | \e , \Ydoz )} \! = \! 
\frac{P(\e | \Xuno \Ydoz ) P(\Xuno \Ydoz )}
     {P(\e | \Xdoz  \Ydoz ) P(\Xdoz \Ydoz)}
\end{displaymath}
The middle equality follows from the fact
that $I(X,Y \mid \vare=\e)$holds for $P$.  $\Box$

\begin{lemma} \label{lem12}
Let $A_1$, $A_2$, $A_3$, $A_4$, $B_1$, $B_2$, $B_3$, and $B_4$
be disjoint sets of variables, and 
$e$ be a single binary variable not contained in any of these sets.
Let $P$ be a probability distribution over the union of these variables.
If the antecedents $I_1$, $I_2$, and $I_3$ of Equation~\ref{pt-bin}
hold for $P$, 
then the following conditions must also hold:
\begin{eqnarray}
I(A_1, \vare \mid \Atwo \Atre \Afor \Bone \Btwo \Btre \Bfor) 
& \Rightarrow &
I(A_1, \vare \mid \Atwo \Atre \Afor B_1  B_2  \Btre \Bfor) \label{lem12a} 
\;\;\;\;  \\
I(B_1, \vare \mid \Aone \Atwo \Atre \Afor \Btwo \Btre \Bfor) 
& \Rightarrow &
I(B_1, \vare \mid A_1 A_2 \Atre \Afor \Btwo \Btre \Bfor) \label{lem12b} \\
I(A_1, \vare \mid \Atwo \Atre \Afor \Bone \Btwo \Btre \Bfor) 
& \Rightarrow &
I(A_1, \vare \mid \Atwo \Atre \Afor B_1  \Btwo B_3 \Bfor) \label{lem13a} \\
I(B_1, \vare \mid \Aone \Atwo \Atre \Afor \Btwo \Btre \Bfor)
& \Rightarrow &
I(B_1, \vare \mid A_1 \Atwo A_3 \Afor \Btwo \Btre \Bfor) \label{lem13b}\\
I(A_1, \vare \mid \Atwo \Atre \Afor \Bone \Btwo \Btre \Bfor)
& \Rightarrow &
I(A_1, \vare \mid \Atwo \Atre A_4 B_1  \Btwo \Btre \Bfor) \label{lem123a} \\
I(B_1, \vare \mid \Aone \Atwo \Atre \Afor \Btwo \Btre \Bfor)
& \Rightarrow &
I(B_1, \vare \mid A_1  \Atwo \Atre \Afor \Btwo \Btre B_4) \label{lem123b}
\end{eqnarray}
where each $\bf A_i'$ and $\bf B_i'$ denote a specific value
for $A_i$ and $B_i$, respectively.
(In words, Equation~\ref{lem12a} states that if $A_1$ and $e$
are conditionally independent for one specific value $\Bone$ of $B_1$
and $\Btwo$ of 
$B_2$, then they are conditionally independent given every
value of $B_1$ and $B_2$, provided the values of
the other variables remain unaltered. 
The other five equations have a similar
interpretation.)
\end{lemma}
\noindent {\bf Proof:} First, we prove Equation~\ref{lem12a}.  
Then we show that the proofs of Equations~\ref{lem12b} 
through \ref{lem13b} are symmetric.  Finally, we will prove
Equations \ref{lem123a} and \ref{lem123b}.
Let $X= A_1 A_2 B_3 B_4$ and $Y=B_1 B_2 A_3 A_4$. 
Then, Lemma~\ref{lemBayes} and $I_2$ yield the following equation:
\begin{eqnarray*}
& & \frac{ P(\e | \Aone \Atwo \Atre \Afor  \Bone \Btwo \Btre \Bfor ) 
P(\Aone \Atwo \Atre \Afor \Bone \Btwo \Btre \Bfor )
}{P(\e | \sAone  \Atwo  \Atre  \Afor \Bone \Btwo  \Btre  \Bfor )   
P(\sAone \Atwo \Atre \Afor \Bone \Btwo \Btre \Bfor )} \\
& & \\
& = & \frac{ P(\e | \Aone \Atwo \Atre \Afor  \sBone \sBtwo \Btre \Bfor ) 
P(\Aone \Atwo \Atre \Afor \sBone \sBtwo \Btre \Bfor )
}{P(\e | \sAone  \Atwo  \Atre  \Afor \sBone \sBtwo  \Btre  \Bfor )   
P(\sAone \Atwo \Atre \Afor \sBone \sBtwo \Btre \Bfor )} \\
\end{eqnarray*}
where $\sAone$, $\sBone$, and $\sBtwo$ are 
arbitrary instances of $A_1$, $B_1$, and $B_2$,
respectively.  Applying $I_1$ and cancelling equal terms yields
\begin{equation}
\label{useit}
\frac{P(\e | \Aone \Atwo \Atre \Afor  \Bone \Btwo \Btre \Bfor)
}{P(\e | \sAone  \Atwo  \Atre  \Afor \Bone \Btwo  \Btre  \Bfor )} =
\frac{ P(\e | \Aone \Atwo \Atre \Afor  \sBone \sBtwo \Btre \Bfor )
}{P(\e | \sAone  \Atwo  \Atre  \Afor \sBone \sBtwo  \Btre  \Bfor )}
\end{equation}
Furthermore, 
$I(A_1, \vare \mid \Atwo \Atre \Afor \Bone \Btwo \Btre \Bfor)$
(the antecedent of Equation~\ref{lem12a}) implies that
\begin{displaymath}
P(\e | \Aone \Atwo \Atre \Afor  \Bone \Btwo \Btre \Bfor) =
P(\e | \sAone \Atwo  \Atre  \Afor \Bone \Btwo  \Btre  \Bfor )
\end{displaymath}
Thus, from Equation~\ref{useit}, it follows that
\begin{equation}
P(\e | \Aone \Atwo \Atre \Afor  \sBone \sBtwo \Btre \Bfor) =
P(\e | \sAone  \Atwo  \Atre  \Afor \sBone \sBtwo  \Btre  \Bfor )
\label{equal1}
\end{equation}
Subtracting each side of Equation~\ref{equal1} from $1$ yields
\begin{equation}
P(\ebar | \Aone \Atwo \Atre \Afor  \sBone \sBtwo \Btre \Bfor) =
P(\ebar | \sAone  \Atwo  \Atre  \Afor \sBone \sBtwo  \Btre  \Bfor )
\label{equal2}
\end{equation}
Thus, $I(A_1, \vare \mid \Atwo \Atre \Afor \sBone \sBtwo  \Btre \Bfor)$
holds for $P$.
Because $\sBone$ and $\sBtwo$
are arbitrary instances,
$I(A_1, \vare \mid \Atwo \Atre \Afor B_1  B_2  \Btre \Bfor)$
also holds for $P$. Thus, Equation~\ref{lem12a} is proved.

Equation~\ref{lem12b} is symmetric with respect to
Equation~\ref{lem12a} by switching the role of $A_1$ with
that of $B_1$ and the role of $A_2$ with that of $B_2$.
Equation~\ref{lem13a} is symmetric with respect to
Equation~\ref{lem12a} by switching the roles of $B_2$ and $B_3$.
Equation~\ref{lem13b} is symmetric with respect to
Equation~\ref{lem12b} by switching the roles of $A_2$ and $A_3$.

Now we prove Equation~\ref{lem123a}.
Equation~\ref{lem123b} is symmetric with respect to
Equation~\ref{lem123a} 
by switching the role of $A_1$ with
that of $B_1$ and the role of $A_4$ with that of $B_4$.

Let $X= A_1 A_2 B_3 B_4$ and $Y=B_1 B_2 A_3 A_4$. 
Applying Lemma~\ref{lemBayes} and $I_2$ and then using $I_1$ to
cancel equal terms, yields the following equation:
\begin{eqnarray}
\lefteqn{
\frac{ P(\e | \Aone \Atwo \Atre \Afor  \Bone \Btwo \Btre \Bfor )
P(\Aone \Atwo \Atre \Afor )
}{P(\e | \sAone  \Atwo  \Atre  \Afor \Bone \Btwo  \Btre  \Bfor ) 
P(\sAone \Atwo \Atre \Afor)}} \nonumber \\ & & \nonumber \\
& = &
\frac{ P(\e | \Aone \Atwo \Atre \sAfor  \sBone \Btwo \Btre \Bfor )
P(\Aone \Atwo \Atre \sAfor)
}{P(\e | \sAone  \Atwo  \Atre  \sAfor  \sBone \Btwo  \Btre  \Bfor )
P(\sAone \Atwo \Atre \sAfor)}
\label{frac-1}
\end{eqnarray}
where $\sAone$, $\sBone$, and $\sAfor$ are arbitrary instances of
$A_1$, $B_1$, and $A_4$, respectively.
Similarly, let
$X= A_1 A_3 B_2 B_4$ and $Y=B_1 B_3 A_2 A_4$. Then,
applying Lemma~\ref{lemBayes} and $I_3$ and using $I_1$ to
cancel equal terms, yields the following equation:
\begin{eqnarray}
\lefteqn{
\frac{ P(\ebar | \Aone \Atwo \Atre \Afor  \Bone \Btwo \Btre \Bfor )
P(\Aone \Atwo \Atre \Afor )
}{P(\ebar | \sAone  \Atwo  \Atre  \Afor \Bone \Btwo  \Btre  \Bfor ) 
P(\sAone \Atwo \Atre \Afor)}} \nonumber\\ & & \nonumber \\
& = &
\frac{ P(\ebar | \Aone \Atwo \Atre \sAfor  \sBone \Btwo \Btre \Bfor )
P(\Aone \Atwo \Atre \sAfor)
}{P(\ebar | \sAone  \Atwo  \Atre  \sAfor  \sBone \Btwo  \Btre  \Bfor
) P(\sAone \Atwo \Atre \sAfor)}
\label{frac-2}
\end{eqnarray}
Now $I(A_1, \vare \mid \Atwo \Atre \Afor \Bone \Btwo \Btre \Bfor)$ 
implies the following two
conditions:
\begin{eqnarray}
& P(\e | \Aone \Atwo \Atre \Afor  \Bone \Btwo \Btre \Bfor ) = 
P(\e | \sAone  \Atwo  \Atre  \Afor \Bone \Btwo  \Btre  \Bfor ) &
\label{cancel1} \\
& P(\ebar | \Aone \Atwo \Atre \Afor  \Bone \Btwo \Btre \Bfor ) = 
P(\ebar | \sAone  \Atwo  \Atre  \Afor \Bone \Btwo  \Btre  \Bfor ) &
\label{cancel2}
\end{eqnarray}
After using Equation~\ref{cancel1} to cancel equal terms in
Equation~\ref{frac-1} and using Equation~\ref{cancel2} to cancel
equal terms in Equation~\ref{frac-2}, we compare
Equations~\ref{frac-1} and \ref{frac-2} and obtain 
\begin{equation}
\frac{ P(\e | \Aone \Atwo \Atre \sAfor  \sBone \Btwo \Btre \Bfor )
}{P(\e | \sAone  \Atwo  \Atre  \sAfor  \sBone \Btwo  \Btre  \Bfor )}
=
\frac{ P(\ebar | \Aone \Atwo \Atre \sAfor  \sBone \Btwo \Btre \Bfor )
}{P(\ebar | \sAone  \Atwo  \Atre  \sAfor  \sBone \Btwo  \Btre  \Bfor )}
\label{frac-3}
\end{equation}
Equation~\ref{frac-3} has the form:
\begin{displaymath}
\frac{x}{y} = \frac{1-x}{1-y}
\end{displaymath}
which yields $x=y$.  

Consequently, we obtain
$I(A_1, \vare \mid \Atwo \Atre \sAfor \sBone  \Btwo \Btre \Bfor)$.
Furthermore, because $\sAfor$ and $\sBone$
are arbitrary instances,
$I(A_1, \vare \mid \Atwo \Atre A_4 B_1  \Btwo \Btre \Bfor)$
holds for $P$.  $\Box$

Next, we prove Theorem~\ref{hardtheorem}.
Let $C = A_2 A_3 A_4$ and $D = B_2 B_3 B_4$.
We will see that $I_1$, $I_2$, and $I_3$ imply 
the following four properties:
\begin{eqnarray}
& & \!\!\!\!\!\!\! I(A_1, \vare \mid C D B_1) \ {\rm or} \ 
I( B_1, \vare \mid C D A_1) 
\label{p1} \\
& & \!\!\!\!\!\!\! I( A_1, \vare \mid C D B_1) \ \Rightarrow  
I(A_1, A_4 \mid A_2 A_3)  \label{p2} \\
& & \!\!\!\!\!\!\! I(A_1, \vare \mid C D B_1) \ \& \ 
I(A_1, A_4 \mid A_2 A_3)
\ \Rightarrow  I(A_1, A_3 \mid A_2)  \label{p3} \\
& & \!\!\!\!\!\!\! I(A_1, \vare \mid C D B_1) \ \& \ 
I(A_1, A_4 \mid A_2 A_3)
\ \& \ I(A_1, A_3 \mid A_2) \ \Rightarrow I(A_1, A_2 \mid \emptyset)
\label{p4}
\end{eqnarray}

First, we prove Equation~\ref{pt-bin}, using these four properties.
Then, we will show that these properties are valid.
From Equation~\ref{p1}, 
there are two symmetric cases to consider.  Without loss of generality,
assume $I(A_1, \vare \mid C D B_1)$ holds. (Otherwise, we switch
the roles of subscripted $A$'s with subscripted $B$'s in
Equations~\ref{p2} through \ref{p4}.)
By a single application of each of Equations~\ref{p2}, \ref{p3}, and
\ref{p4}, the following independence statements are proved to 
hold for $P$:
\begin{displaymath}
I(A_1, A_2 \mid \emptyset), \ I(A_1, A_3 \mid A_2), \ 
I(A_1, A_4 \mid A_2 A_3)
\end{displaymath}
These three statements yield 
$I( A_1 ,  A_2 A_3 A_4 \mid \emptyset)$ ($\equiv I_4$)
by two applications of contraction.
Consider Equation~\ref{pt-bin}.
The statement
$I( A_1 A_2 A_3 A_4,  B_1 B_2 B_3 B_4 \mid \emptyset)$ (i.e., $I_1$)
implies $I( A_1 , B_1 B_2 B_3 B_4 \mid A_2 A_3 A_4)$
using weak union, which together 
with $I_4$ imply using
contraction 
$I( A_1 , A_2 A_3 A_4 B_1 B_2 B_3 B_4 \mid \emptyset)$.
This statement together with the statement
$I( A_1 , \vare \mid C B_1 D)$
imply, using contraction, the statement
$I( A_1 , \vare A_2 A_3 A_4 B_1 B_2 B_3 B_4 \mid \emptyset)$,
thus completing the proof.

It remains to prove Equations~\ref{p1} through
\ref{p4}.  First, we prove Equation~\ref{p1}.
Let $\Auno$, $\Adoz$, $\Buno$, $\Bdoz$,
$\starC$, and $\starD$ be arbitrary instances of $A_1$, $B_1$,
$C$, and $D$, respectively. Let $X=AC$ and $Y=BD$.
Then, Lemma \ref{lemBayes} and $I_2$
yield the following equation:
\begin{equation}
\frac{P(\e | \Auno  \starC \starD  \Buno) P(\Auno \starC \starD \Buno)}
{ P(\e | \Adoz \starC \starD  \Buno)  P(\Adoz \starC \starD \Buno )} = 
\frac{P(\e | \Auno  \starC \starD  \Bdoz) P(\Auno \starC \starD \Bdoz )}
{P(\e | \Adoz \starC \starD  \Bdoz ) P(\Adoz \starC \starD \Bdoz)}
\label{frac1}
\end{equation}
From $I_1$, we obtain $P(ACDB) = P(AC)P(DB)$.  
Consequently, Equation~\ref{frac1} yields
\begin{equation}
\frac{P(\e | \Auno \Buno  \starC \starD)}
{P(\e | \Adoz \Buno \starC \starD )} = 
\frac{P(\e | \Auno \Bdoz \starC \starD)}
{P(\e | \Adoz \Bdoz \starC \starD)} \label{frac2}
\end{equation}
Equation~\ref{frac2} has the following algebraic form, where
subscripted $X$s replace the corresponding terms:
\begin{equation}
\frac{ X_{\Auno\Buno}}{X_{\Adoz\Buno} } = 
\frac{ X_{\Auno\Bdoz}  }{ X_{\Adoz\Bdoz}} \label{fracX}
\end{equation}
Using Lemma \ref{lemBayes} and $I_3$, we obtain a
relationship similar to Equation~\ref{frac2}, where the
only change is that $\e$ is replaced with $\ebar$:
\begin{equation}
\frac{P(\ebar | \Auno  \Buno \starC \starD)}
{ P( \ebar | \Adoz \Buno  \starC \starD )} = 
\frac{P(\ebar | \Auno  \Bdoz \starC \starD)}
{P(\ebar | \Adoz \Bdoz \starC \starD)} \label{frac2bar}
\end{equation}
We rewrite Equation~\ref{frac2bar} in terms of $X$s, 
and then use Equation~\ref{fracX} to obtain
\begin{equation}
\frac{1- X_{\Auno\Buno}}{1- X_{\Adoz\Buno}} = 
\frac{ 1 - k  X_{\Auno\Buno}}{1- k  X_{\Adoz\Buno}} \label{fracXbar}
\end{equation}
where $k =\frac{ X_{\Auno\Bdoz}}{ X_{\Auno\Buno}}$.  
Equation~\ref{fracXbar} implies that either
$X_{\Auno\Bdoz} = X_{\Auno\Buno}$ (i.e., $k = 1$) or
$X_{\Auno\Buno} = X_{\Adoz\Buno}$.
Because the choice of instances for $A_1$ and $B_1$ is arbitrary,
at least one of the following two sequences of equalities must hold:

\begin{itemize}
\item
For every instance {$\bf B$} of $B_1, \ X_{\bf A^1 B } = 
X_{\bf A^2 B}  = ... = X_{\bf A^m B}$

\item
For every instance {$\bf A$} of $A_1, \ X_{\bf A B^1 } = 
X_{\bf A B^2 } = ... = X_{\bf A B^n }$
\end{itemize}
where $\bf A^1 ,\ldots, A^m$ are the instances of $A_1$ and
$\bf B^1 ,\ldots, B^n$ are the instances of $B_1$.

Thus, by definition of the $X$s, we obtain
\begin{equation}
\forall \starC \starD \ {\rm instances \ of} \ CD \;\;\;
[ I(\vare,  A_1 \mid \starC \starD B_1 ) \ {\rm or} \ 
I(\vare,  B_1 \mid \starC \starD A_1)] \label{pivot}
\end{equation}
On the other hand, Equation~\ref{p1}, which we are now proving, states
\begin{equation}
[\forall \starC \starD \; I(\vare,  A_1 \mid \starC \starD B_1)] \;\;
{\rm or} \; \;
[\forall \starC \starD \ I(\vare, B_1 \mid  \starC \starD A_1)] \label{goal}
\end{equation}
which is stronger than Equation~\ref{pivot}.
Equation~\ref{p1} can also be written as follows:
\begin{equation}
\neg I(B_1, \vare \mid C D A_1) \Rightarrow I(A_1, \vare \mid C D B_1)  
\label{chain1}
\end{equation}

We prove Equation~\ref{chain1}.  
The statement $\neg I(B_1, \vare \mid C D A_1)$ implies that
there exists instances 
$\Aone$, $\Atwo$, $\Atre$, $\Afor$, 
$\Bone$, $\Btwo$, $\Btre$, $\Bfor$, 
and $\eone$ of
$A_1$, $A_2$, $A_3$, $A_4$, 
$B_1$, $B_2$, $B_2$, $B_3$, $B_4$, and $\vare$, respectively,
such that
\begin{equation}
\neg I(\Bone,\eone \mid \Aone\Atwo\Atre\Afor \Btwo\Btre\Bfor)
\end{equation}
Hence,
\begin{equation}
\neg I(B_1, e \mid A_1 A_2 \Atre\Afor \Btwo\Btre\Bfor) \label{ch1-1}
\end{equation}
From Lemma~\ref{lem12} (contrapositive form of Equation~\ref{lem12b}),
Equation~\ref{ch1-1} implies
\begin{equation}
\neg I(B_1, e \mid \sAone \sAtwo \Atre \Afor \Btwo\Btre\Bfor) 
\end{equation}
where $\sAone$ and $\sAtwo$ are arbitrary instances of $A_1$
and $A_2$, respectively.
Hence, in particular, if $\sAone=\Aone$, we have
\begin{equation}
\neg I(B_1, e \mid \Aone \sAtwo \Atre \Afor \Btwo\Btre\Bfor )
\label{ch1-2}
\end{equation}
Similarly, from Lemma~\ref{lem12} (Equation~\ref{lem13b}), 
Equation~\ref{ch1-2} implies
\begin{equation}
\neg I(B_1, e \mid \sAone \sAtwo \sAtre \Afor \Btwo\Btre\Bfor) \label{ch1-3}
\end{equation}
where $\sAtre$ is an arbitrary instance
of $A_3$.
Also, from Lemma~\ref{lem12} (Equation~\ref{lem123b}), 
Equation~\ref{ch1-3} implies
\begin{equation}
\neg I(B_1, e \mid \sAone \sAtwo \sAtre \Afor \Btwo\Btre \sBfor) \label{ch1-4}
\end{equation}
where $\sBfor$ is an arbitrary instance of $B_4$.
Examine Equation~\ref{pivot}.
Equation~\ref{ch1-4} states that the second
disjunct cannot be true for
every instance of $A_1 A_2 A_3 B_1 B_4$ and $e$.
Hence, for each of these instances the
other disjunct must hold.  That is,
\begin{equation}
\forall \sAone \sAtwo \sAtre \sBone \sBfor \stare \; 
I(\sAone, \stare \mid \sBone\sAtwo\sAtre\Afor \Btwo\Btre \sBfor)
\end{equation}
or, equivalently,
\begin{equation}
I(A_1, e \mid B_1 A_2 A_3 \Afor \Btwo\Btre B_4) 
\label{ch1-5}
\end{equation}
Applying Equation~\ref{ch1-5} to
Equation~\ref{lem12a} yields,
\begin{equation}
I(A_1, e \mid B_1 A_2 A_3 \Afor B_2 \Btre B_4) 
\label{ch1-6}
\end{equation}
Similarly, applying Equation~\ref{ch1-6} to
Equations~\ref{lem13a}, and \ref{lem123a}
yields the statement
\begin{equation}
I(A_1, \vare \mid A_2 A_3 A_4 B_1 B_2 B_3 B_4)
\label{ch1-7}
\end{equation}
which is the desired consequence of Equation~\ref{chain1}.
Thus, we
have proved Equation~\ref{p1}.

Next, we show that Equation~\ref{p2} must hold.
Lemma~\ref{lemBayes} and $I_2$ yield the following equation: 
\begin{eqnarray*}
\lefteqn{
\frac{ P(\e | \tsAone \sAtwo \sAtre \tsAfor  
\sBone \sBtwo \sBtre \sBfor ) 
P(\tsAone \sAtwo \sAtre \tsAfor \sBone \sBtwo \sBtre \sBfor )}
{ P(\e | \tAone  \sAtwo  \sAtre  \tsAfor  
\sBone \sBtwo  \sBtre  \sBfor )   
P(\tAone  \sAtwo  \sAtre \tsAfor \sBone  \sBtwo  \sBtre  \sBfor )   
}} \\
& & \\
& = & \frac{ P(\e | \tsAone  \sAtwo  \sAtre \tAfor  
\sBone  \sBtwo  \sBtre  \sBfor )   
P(\tsAone \sAtwo  \sAtre \tAfor \sBone  \sBtwo  \sBtre  \sBfor )}
{ P(\e | \tAone  \sAtwo  \sAtre \tAfor   
\sBone  \sBtwo  \sBtre  \sBfor )   
P(\tAone \sAtwo  \sAtre \tAfor \sBone  \sBtwo  \sBtre  \sBfor )}.
\end{eqnarray*}
Incorporating $I_1$ and
$I( e , A_1 \mid A_2 A_3 A_4 B_1 B_2 B_3 B_4 )$ 
(Equation~\ref{ch1-7}),
and cancelling some equal terms yields
\begin{eqnarray*}
\lefteqn{
\frac{ 
P(\tsAfor |  \tsAone \sAtwo \sAtre  )  
P( \tsAone \sAtwo \sAtre  )  
P( \sBone \sBtwo \sBtre \sBfor )}
{
P(\tsAfor |  \tAone  \sAtwo  \sAtre )  
P( \tAone  \sAtwo  \sAtre )  
P( \sBone  \sBtwo  \sBtre  \sBfor )   
}} \\
& & \\
& = & \frac{
P(\tAfor |  \tsAone \sAtwo  \sAtre  )  
P( \tsAone \sAtwo  \sAtre  )  
P( \sBone  \sBtwo  \sBtre  \sBfor )}
{
P(\tAfor |   \tAone \sAtwo  \sAtre  )  
P( \tAone \sAtwo  \sAtre    )  
P( \sBone  \sBtwo  \sBtre  \sBfor )}.
\end{eqnarray*}
Further cancelation of equal terms yields

\begin{displaymath}
\frac{ P( \tsAfor |  \tsAone  \sAtwo  \sAtre) }
{ P( \tAfor |  \tsAone  \sAtwo  \sAtre) } = 
\frac{ P( \tsAfor |  \tAone  \sAtwo \sAtre) }
{ P( \tAfor |  \tAone  \sAtwo  \sAtre  ) . }
\end{displaymath}
Thus,
$P( \tsAfor |  \tsAone \sAtwo  \sAtre ) =  P( \tsAfor | \tAone 
\sAtwo  \sAtre )$ 
for every instance $\tsAone , \tAone$, and $\tsAfor$.
That is, $I( A_4 , A_1 \mid  \sAtwo  \sAtre )$ holds. 
Because $ \sAtwo $ and $ \sAtre $ 
are arbitrary instances, $I( A_4 , A_1 \mid A_2 A_3 )$ follows. 

Next, we show that Equation~\ref{p3} must hold.
Lemma \ref{lemBayes} and $I_2$ 
yield the following equation:
\begin{eqnarray*}
\lefteqn{\frac{ P(\e | \tsAone \sAtwo \tsAtre \sAfor
\sBone \sBtwo \sBtre \sBfor )
P(\tsAone \sAtwo \tsAtre \sAfor \sBone \sBtwo \sBtre \sBfor )    
}{ 
P(\e | \tAone \sAtwo \tsAtre \sAfor \sBone \sBtwo \sBtre \sBfor )
P(\tAone \sAtwo \tsAtre \sAfor \sBone \sBtwo \sBtre \sBfor )}} \\
& & \\
& = & 
\frac{ P(\e | \tsAone \sAtwo \tAtre \sAfor \sBone \sBtwo \sBtre \sBfor )
P(\tsAone \sAtwo \tAtre \sAfor \sBone \sBtwo \sBtre \sBfor )
}{
P(\e | \tAone \sAtwo \tAtre \sAfor
\sBone \sBtwo \sBtre \sBfor )
P(\tAone \sAtwo \tAtre \sAfor \sBone \sBtwo \sBtre \sBfor )}
\end{eqnarray*}

Incorporating 
$I_1$, $I(A_1 , A_4 \mid A_2 A_3 )$, and
$I( e , A_1 \mid A_2 A_3 A_4 B_1 B_2 B_3 B_4 )$ 
and cancelling some equal terms yields

\begin{eqnarray*}
\lefteqn{\frac{
P(\sAfor |  \sAtwo \tsAtre )  
P(\tsAtre |  \tsAone \sAtwo )  
P( \tsAone \sAtwo  )  
P( \sBone \sBtwo \sBtre \sBfor )    
}{ 
P(\sAfor |  \sAtwo \tsAtre )  
P(\tsAtre |  \tAone \sAtwo )  
P( \tAone \sAtwo  )  
P( \sBone \sBtwo \sBtre \sBfor )}} \\
& & \\
& = & \frac{ 
P(\sAfor |  \sAtwo \tAtre )  
P(\tAtre |  \tsAone \sAtwo )  
P( \tsAone \sAtwo  )  
P( \sBone \sBtwo \sBtre \sBfor )
}{
P(\sAfor |  \sAtwo \tAtre )  
P(\tAtre |  \tAone \sAtwo )  
P( \tAone \sAtwo  )  
P( \sBone \sBtwo \sBtre \sBfor )}
\end{eqnarray*}
Further cancelation of equal terms yields

\begin{displaymath}
\frac{P( \tsAtre |  \tsAone  \sAtwo)}{ P( \tAtre |  \tsAone  \sAtwo  )} = 
\frac{P( \tsAtre |  \tAone  \sAtwo )}{ P( \tAtre |  \tAone  \sAtwo )}
\end{displaymath}
Thus,
$P( \tsAtre |  \tsAone \sAtwo ) =  P( \tsAtre | \tAone  \sAtwo )$ 
for every instance $\tsAone , \tAone$ and $\tsAtre$.  That is,
$I( A_3 , A_1 \mid  \sAtwo )$ holds. 
Because $ \sAtwo $ 
is an arbitrary instance, 
$I( A_3 , A_1 \mid A_2 )$ follows. 

Finally, we must show that Equation~\ref{p4} holds.
Lemma \ref{lemBayes} and $I_3$
yield the following equation:
\begin{eqnarray*}
\lefteqn{\frac{P( \ebar | \tsAone \tsAtwo \sAtre \sAfor
\sBone \sBtwo \sBtre \sBfor )
P(\tsAone \tsAtwo \sAtre \sAfor \sBone \sBtwo \sBtre \sBfor )
}{
P( \ebar | \tAone \tsAtwo \sAtre \sAfor
\sBone \sBtwo \sBtre \sBfor )
P(\tAone \tsAtwo \sAtre \sAfor \sBone \sBtwo \sBtre \sBfor )}} \\
& & \\
& = &
\frac{ P(\ebar | \tsAone \tAtwo \sAtre \sAfor
\sBone \sBtwo \sBtre \sBfor)
P(\tsAone \tAtwo \sAtre \sAfor \sBone \sBtwo \sBtre \sBfor )
}{
P( \ebar | \tAone \tAtwo \sAtre \sAfor
\sBone \sBtwo \sBtre \sBfor)
P(\tAone \tAtwo \sAtre \sAfor \sBone \sBtwo \sBtre \sBfor)}
\end{eqnarray*}

Incorporating $I( e, A_1 \mid A_2 A_3 A_4 B_1 B_2 B_3 B_4 )$,
$I_1$, $I_3$, $I(A_1 , A_4 \mid A_2 A_3 )$, and
$I(A_1 , A_3 \mid A_2 )$
and cancelling some equal terms yields
\begin{eqnarray*}
\lefteqn{\frac{
P(\sAfor |  \tsAtwo * \sAtre )
P(\sAtre |  \tsAone \tsAtwo )
P(\tsAtwo |  \tsAone )
P(\tsAone )  
P(\sBone \sBtwo \sBtre \sBfor )
}{
P(\sAfor |  \tsAtwo * \sAtre )  
P(\sAtre |  \tAone \tsAtwo  )  
P(\tsAtwo |  \tAone )  
P(\tAone )  
P(\sBone \sBtwo \sBtre \sBfor )}} \\
& & \\
& = &
\frac{
P(\sAfor |  \tAtwo * \sAtre )
P(\sAtre |  \tsAone \tAtwo  )
P(\tAtwo |  \tsAone )
P(\tsAone )
P(\sBone \sBtwo \sBtre \sBfor )
}{
P(\sAfor |  \tAtwo * \sAtre)  
P(\sAtre |  \tAone \tAtwo )
P(\tAtwo |  \tAone )
P(\tAone )
P(\sBone \sBtwo \sBtre \sBfor)}
\end{eqnarray*}
Further cancelation of equal terms yields

\begin{displaymath}
\frac{P( \tsAtwo | \tsAone)}{ P( \tAtwo |  \tsAone)} = 
\frac{P( \tsAtwo | \tAone)}{ P( \tAtwo |  \tAone  ) }
\end{displaymath}
Thus,
$P( \tsAtwo |  \tsAone ) =  P( \tsAtwo | \tAone )$ 
for every instance $\tsAone , \tAone$ and $ \tsAtwo$.
That is, $I( A_2 , A_1 \mid \emptyset )$ holds.  $\Box$

We conjecture that Theorem~\ref{hardtheorem}
holds also for binary distributions that are not
strictly positive.

\section*{Appendix B: Regular Gaussian distributions}

We show that Equation~\ref{pt-bin} holds for
regular Gaussian distributions.  Our proof is based
on three properties of 
regular Gaussian distributions:
\begin{eqnarray}
& & I(X, Y \mid Z)\ \&\ I(X, W \mid Z)  \Rightarrow    
I(X, YW \mid Z) \label{composition} \\
& & I(X, Y \mid {\bf Z})  \Rightarrow  I(X, Y \mid Z) \label{unification} \\
& & I(X, Y \mid \emptyset) \ \& \ I(X, Y \mid e) \Rightarrow
I(X, e \mid \emptyset) \ {\rm or} \ I(e, Y \mid \emptyset) \label{marginal_wt}
\end{eqnarray}
where $e$ is a single variable not contained in $XY$.

The first two properties can be verified
trivially from the definition of 
regular Gaussian distributions, whereas the third requires
some algebra on the determinant of a covariance 
matrix.  These considerations are left to the reader.

Equation~\ref{unification} is an interesting property.
It states that, for Gaussian distributions, if $X$ and $Y$
are conditionally independent given a specific value $\bf Z$ of $Z$, 
then $X$ and $Y$ are conditionally independent for every value of $Z$.
The truth of this property rests on the fact
that whether or not $I(X, Y \mid Z)$ holds for
a Gaussian distribution is determined solely by its covariance
matrix, which does not depend on the values given to $X, Y$ and $Z$.

The three statements in the antecedents of 
Equation~\ref{pt-bin} are 
\begin{displaymath}
I( A_1 A_2 A_3 A_4 ,  B_1 B_2 B_3 B_4 \mid \emptyset )
\end{displaymath}  
\begin{displaymath}
I( A_1 A_2 B_3 B_4 , B_1 B_2 A_3 A_4 \mid \eone)   
\end{displaymath}  
\begin{displaymath}
I( A_1 A_3 B_2 B_4 , B_1 B_3 A_2 A_4 \mid \etwo)
\end{displaymath}
Applying Equation~\ref{unification} yields the following statements,
where $\eone$ and $\etwo$ are replaced with their variable name $e$:  
\\
\hspace*{4cm}
$I( A_1 A_2 A_3 A_4 , B_1 B_2 B_3 B_4 \mid \emptyset)$ \hfill ($I_1$)\\
\hspace*{4cm}
$I( A_1 A_2 B_3 B_4 , B_1 B_2 A_3 A_4 \mid e)$ \hfill ($I_2$)\\
\hspace*{4cm}
$I( A_1 A_3 B_2 B_4 , B_1 B_3 A_2 A_4 \mid e)$ \hfill ($I_3$)\\
The following three Equations are also needed for the proof:
\begin{eqnarray}
\lefteqn{I( A_1 A_2 , B_1 B_2 \mid \emptyset) \ \& \ 
I( A_1 A_2 , B_1 B_2 \mid e)  \Rightarrow  \nonumber}\\
& & \mbox{\hspace{3cm}} 
I( A_1 A_2 , e \mid \emptyset) \ {\rm or} \ 
I(e, B_1 B_2 \mid \emptyset)
\label{6.a} \\ 
& & \mbox{\hspace{-.71cm}} 
I( A_1 A_2 , A_3 A_4 \mid e)  \ \& \  I( A_1 A_2 , e \mid \emptyset)  
\Rightarrow  I( A_1 A_2 , e A_3 A_4 \mid \emptyset)
\label{6.b} \\
\lefteqn{I( A_1 A_2 , B_1 B_2 B_3 B_4 \mid \emptyset)  \ \& \ 
I( A_1 A_2 , e A_3 A_4 \mid \emptyset)  
\Rightarrow \nonumber} \\
& & \mbox{\hspace{3cm}} 
I( A_1 A_2 , e A_3 A_4 B_1 B_2 B_3 B_4 \mid \emptyset) \label{6.c}
\end{eqnarray}
Equations~\ref{6.a}, \ref{6.b}, and \ref{6.c} are special
cases of Equations 
\ref{marginal_wt}, \ref{contraction}, and
\ref{composition}, respectively.
Next, we prove the theorem by showing that the right hand side of 
Equation~\ref{pt-bin}, namely
\begin{equation}
I( A_1 , e A_2 A_3 A_4 B_1 B_2 B_3 B_4 \mid \emptyset) \ {\rm or} \  
I( B_1 , e A_1 A_2 A_3 A_4 B_2 B_3 B_4 \mid \emptyset)
\label{bravo}
\end{equation}
follows from $I_1$, $I_2$, and $I_3$. 

First, note that the two antecedents of 
Equation~\ref{6.a} follow, using decomposition
on $I_1$ and $I_2$, respectively.  Thus, Equation~\ref{6.a}
yields
\begin{equation}
I( A_1 A_2 ,e \mid \emptyset) \ {\rm or} \  
I( B_1 B_2 ,e \mid \emptyset) \label{disjunct1}
\end{equation}

Assume the first disjunct holds. This is the second
antecedent of Equation~\ref{6.b}.
The first antecedent of Equation~\ref{6.b} 
follows from $I_2$, using decomposition.  
Thus, Equation~\ref{6.b} yields 
$I( A_1 A_2 , e A_3 A_4 \mid \emptyset)$,
which is the second antecedent of Equation~\ref{6.c}.
The first antecedent of Equation~\ref{6.c} follows form $I_1$ using
decomposition.
Thus, Equation~\ref{6.c} yields \\
\hspace*{4cm}
$ I( A_1 A_2 ,  e A_3 A_4 B_1 B_2 B_3 B_4 \mid \emptyset)$ 
\hfill $(J_1)$ \\
Now assume the second disjunct of Equation~\ref{disjunct1} holds.  
A similar derivation, where
the roles of $A$s and $B$s are switched in
Equations~\ref{6.a} through \ref{6.c}, yields\\
\hspace*{4cm}
$I( B_1 B_2 , e B_3 B_4 A_1 A_2 A_3 A_4 \mid \emptyset)$
\hfill $(J_2)$ \\
Consequently, we have shown that $I_1$ and $I_2$ 
imply $J_1 \;{\rm or}\; J_2$.

Similarly, 
by switching the role of $A_2$ with that of $A_3$
and the role of $B_2$ with that of $B_3$, and using $I_3$
instead of $I_2$,
we obtain\\
\hspace*{4cm}
$I( A_1 A_3 , e A_2 A_4 B_1 B_2 B_3 B_4 \mid \emptyset)$ 
\hfill  $(J_3 )$\\
\hspace*{6cm} or \hfill \\
\hspace*{4cm}
$I( B_1 B_3 , e B_2 B_4 A_1 A_2 A_3 A_4 \mid \emptyset)$
\hfill  $(J_4 )$\\

Thus,
there are four cases to consider, by choosing one statement of
each of the two disjunctions above.
The four cases ar $J_1$ and $J_3$, $J_2$ and $J_4$, $J_1$
and $J_4$, and $J_2$ and $J_3$. 

If $J_1$ and $J_3$ hold, then
$I( A_1 , e A_3 A_4 B_1 B_2 B_3 B_4 \mid \emptyset)$
follows from $J_1$ and $I( A_1 , A_2 \mid \emptyset)$ 
follows from $J_3$, using decomposition.  
Together, using Equation~\ref{6.a}, the statement
$I( A_1 , e A_2 A_3 A_4 B_1 B_2 B_3 B_4 \mid \emptyset)$
is implied.  Similarly, when $J_2$ and $J_4$ hold, 
$I( B_1 , e A_1 A_2 A_3 A_4 B_2 B_3 B_4 \mid \emptyset)$
must hold.  If $J_1$ and $J_4$ hold, then, using decomposition on $J_4$,
$I( B_3 , e \mid \emptyset)$ is obtained.  The
statement $I( B_3 , B_1 \mid e )$
is implied from $I_2$ using decomposition.  
Together, the two statements yield, 
using contraction and decomposition, 
$I( B_3 , B_1 \mid \emptyset )$.  
This statement combined with
$I( B_1 , e B_2 B_4 A_1 A_2 A_3 A_4 \mid \emptyset)$,
which follows from $J_4$ using decomposition, imply,
using symmetry and Equation~\ref{6.a}, that the statement
$I( B_1 , e A_1 A_2 A_3 A_4 B_2 B_3 B_4 \mid \emptyset)$
holds.  
The case where $J_2 $ and $J_3$ hold is symmetric to
the case where $J_1$ and $J_4$ hold 
(by switching the roles of the $A$'s with those of the $B$'s), 
thus yielding
$I( A_1 , e A_2 A_3 A_4 B_1 B_2 B_3 B_4 \mid \emptyset)$.  

We have shown that for each of the four possible cases at
least one of the disjuncts of Equation~\ref{bravo} is implied.
Thus, Equation~\ref{pt-bin} holds.
$\Box$
\end{document}